\documentclass[10pt]{article}
\usepackage[utf8]{inputenc}
\PassOptionsToPackage{hidelinks}{hyperref}
\usepackage{preprint}
\usepackage{orcidlink}
\usepackage{amsmath}
\usepackage{amssymb}
\usepackage{graphicx}
\graphicspath{{figures/}}
\usepackage{float}
\usepackage{booktabs}
\usepackage{hyperref}
\newcommand{\methods}[2]{\hyperref[#1]{Methods, `#2'}}
\newcommand{\edfig}[1]{\hyperref[fig:ED#1]{Extended Data Fig.~\ref*{fig:ED#1}}}
\usepackage{caption}
\DeclareCaptionLabelSeparator{bar}{ | }
\usepackage[backend=biber, style=numeric-comp, sorting=none, autocite=superscript]{biblatex}
\makeatletter
\def\fps@figure{tbp}
\makeatother


\title{Extremely coarse learning objectives induce human-aligned representations in ai vision models}

\author{
    \textbf{Yash Mehta$^1$}~\orcidlink{0000-0002-9610-7077},
    \textbf{Michael F. Bonner$^1$}~\orcidlink{0000-0002-4992-674X}\\
    $^1$Department of Cognitive Science, Johns Hopkins University\\
    \texttt{\href{mailto:ymehta3@jhu.edu}{ymehta3@jhu.edu}, \href{mailto:mfbonner@jhu.edu}{mfbonner@jhu.edu}}
}

\date{}

\begin{document}

\maketitle

\begin{abstract}
    \textbf{Artificial neural networks trained on visual tasks develop internal representations resembling those of the primate visual system, a discovery that has guided a decade of computational neuroscience. Research on building brain-aligned models has progressively embraced finer-grained learning objectives, from object classification to contrastive self-supervised objectives that maximize distinctions among individual images. Yet the effect of learning-signal granularity on brain alignment remains largely unexamined. Here we systematically investigate how the granularity of a learning signal shapes representational alignment with human vision. We parametrically vary the number of training classes using a data-driven approach that partitions a set of training images into different numbers of categories via PCA-based splits of pretrained embeddings. We train hundreds of neural networks across convolutional and transformer architectures on these coarse classification tasks and compare their representations with human fMRI responses, macaque electrophysiology recordings, and human behavior. We find that networks trained to distinguish as few as eight broad categories learn representations that match or exceed the neural alignment of models distinguishing 1,000 classes. Even more strikingly, these coarsely trained networks align more closely with human perceptual similarity judgments than all other models evaluated, including networks trained with fine-grained supervision or self-supervision as well as leading large-scale vision models. These results demonstrate that human-like visual representations can emerge from surprisingly simple learning objectives, reframing what learning signals vision may require and opening a path toward building AI systems that are more aligned with human perception.}
\end{abstract}

\section{Introduction}\label{introduction}

Deep neural networks trained to classify images into many different categories develop internal representations that are well-aligned with neural responses across the primate ventral visual stream\autocite{cadieu2014,yamins2014,khalighrazavi2014}. These results motivated the use of neural networks as models of biological vision. However, a crucial question remains largely unexamined: how complex does the optimization objective need to be to yield brain-like representations? Answering this question has broad implications for understanding the fundamental factors governing alignment across biological and artificial vision and for understanding what kind of supervisory signals suffice for learning a brain-like visual hierarchy.

Recent models of biological vision have used both thousand-way classification and self-supervised objectives that distinguish individual images\autocite{konkle2022,margalit2024,doerig2025}. Yet there is little evidence that the brain has access to comparably fine-grained supervision during learning. Categorization in early development proceeds along coarse distinctions (animate versus inanimate, natural versus artificial) well before children can identify objects at subordinate levels\autocite{rosch1976}, and these broad categorical boundaries are among the most robust organizing principles of high-level visual cortex\autocite{kriegeskorte2008,grillspector2014,konkle2013}. We therefore tested whether training on these broad distinctions could produce representations aligned with biological vision.

To test this possibility, we require a method that can vary the number of training classes in a principled and controlled manner. We developed a data-driven, modality-agnostic procedure that derives coarse category labels directly from the statistical structure of visual inputs, without manual annotation. We applied this procedure to ImageNet\autocite{deng2009} to defines classification tasks ranging from 2 to 64 categories, all much coarser than the standard 1,000 ImageNet classes. We then trained hundreds of networks, varying the number of classes while holding all other aspects of the training setup fixed.

We find that networks trained on coarse categorical distinctions, as few as eight broad classes, develop representations that match or exceed those of fine-grained supervised models in their alignment with neural responses in both human fMRI and macaque electrophysiology recordings. Remarkably, these coarse-trained networks also surpass all tested models, including widely used large-scale models such as CLIP\autocite{radford2021} and DINOv3\autocite{simeoni2025}, in their alignment with human perceptual similarity judgments, a result that holds robustly across architectures and training regimes. Together, these findings reveal that human-aligned visual representations can emerge through optimization on surprisingly coarse categorization objectives. 

\section{Coarse-graining the input space}\label{sec:coarse-graining}

We sought to create a supervised learning framework in which we could systematically vary the number of training classes. In typical supervised tasks, image labels are provided by human annotators, thus leveraging knowledge about how people categorize images. For our supervised tasks, our goal was to create a hierarchy of image labels based on the dominant axes of variation in a learned representation of natural images. We therefore took a different approach than typical supervised learning. Rather than collecting a new set of human annotations or using a predefined semantic hierarchy (e.g., WordNet\autocite{miller1995}), we instead leveraged knowledge baked into the internal representations of pretrained neural networks. The idea is to extract latent information about the organization of high-level image representations that a network has already learned from optimization on a pretraining task (e.g., ImageNet classification) and then use this latent information to create new sets of image labels with different numbers of classes. This approach has connections to network distillation techniques in the sense that we use a teacher network to create training objectives for new networks\autocite{hinton2015}. However, in our work, the goal is not to create smaller networks that maintain good performance on the original task; the goal is to create new task objectives using coarser category labels. 

\begin{figure}
\centering
\includegraphics[width=\linewidth,keepaspectratio]{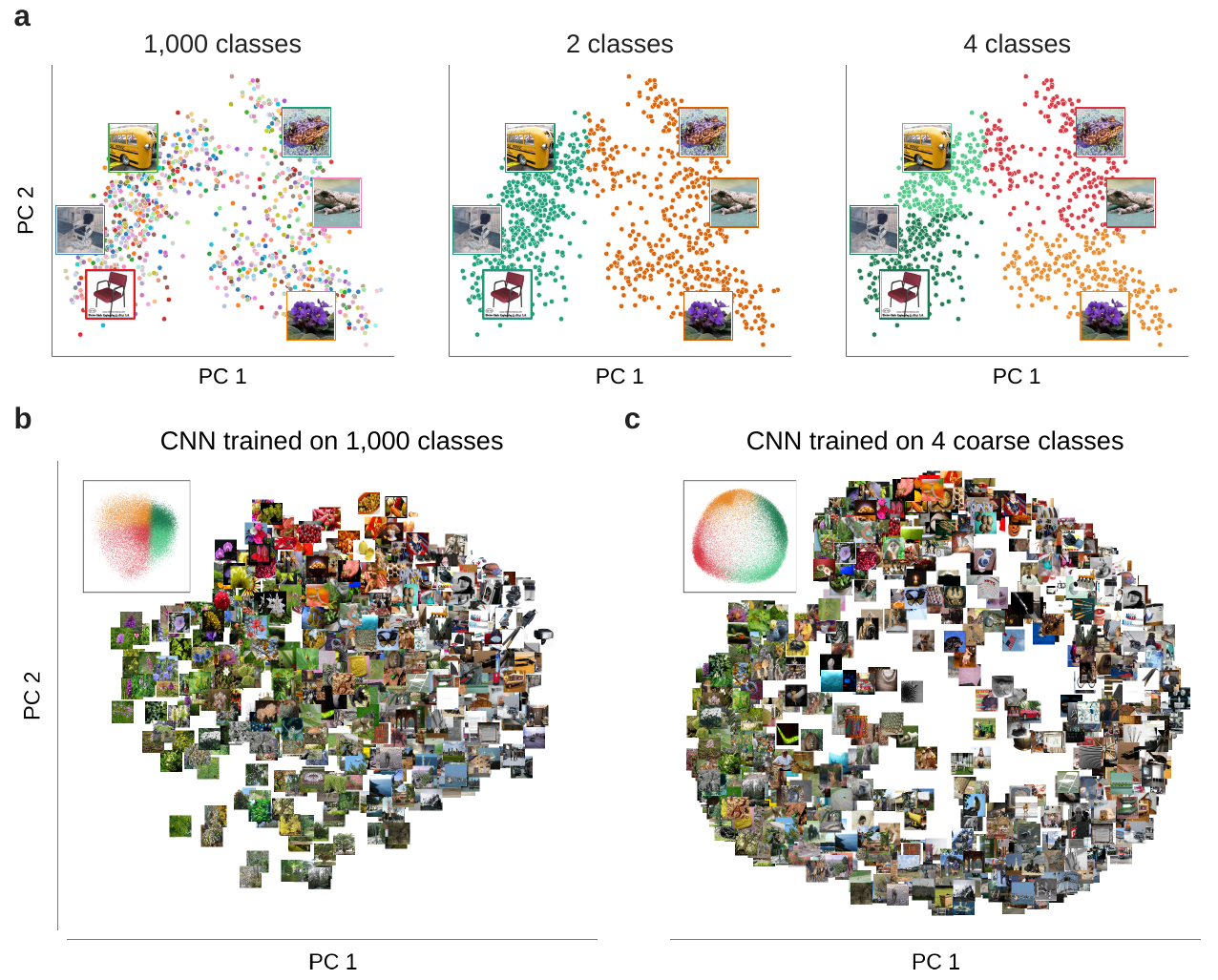}
\caption{\textbf{Training on coarse labels produces a more categorical image representation.}
\textbf{a}, Construction of coarse labels from a pretrained network's image representations. Images are divided at the global median along the first principal component (PC1) to form two classes; adding a split along PC2 produces four classes. The same 1,000 ImageNet images, one per original category, are shown in each panel, colored by their original labels (left), two coarse labels (middle) or four coarse labels (right). Image positions remain fixed, and thumbnail borders indicate the labels assigned to example images.
\textbf{b,c}, Image representations in AlexNet pretrained on ImageNet's 1,000 categories (\textbf{b}) and an AlexNet-style network trained from scratch on four coarse classes derived from the pretrained model (\textbf{c}). Activations from the first fully connected layer (FC1) for 50,000 ImageNet images are projected onto the first two principal components, computed separately for each network. Thumbnails show the same random sample of 1,000 images; insets show all 50,000 images colored by their four-class labels. The coarse classes overlap in the pretrained network's projection but occupy more distinct arcs of a ring in the coarse-trained network's projection.}\label{fig:labels}
\end{figure}

To accomplish this, we extract feature representations of all ImageNet training images from a pretrained source network and compute the principal components of this representation space. The first principal component, capturing the axis of greatest variance, can be used to define a binary partition of the image set via a median split. Each subsequent principal component adds a further median split, so that an image's label is defined by its joint pattern of signs across the first \textit{n} components, yielding $2^n$ categories after \textit{n} splits (Fig.~\ref{fig:labels}a; \methods{sec:pc-partitions}{Principal component partitions}). PCA is fitted once on the full image set and the median for each component is computed across all images; neither is recomputed within subgroups. The resulting partitions are nested, since removing the final split recovers the preceding coarser partition. The pretrained source model serves only to assign a discrete categorical label to each image from a set of K coarse categories, thus reducing the supervisory signal to at most $\log_2(K)$ bits per image. The resulting labels group many visually diverse images into a small number of broad classes (with hundreds of thousands of diverse images per class).

In our main analyses, we present results for coarse-supervised networks trained on classes derived from the principal components of CLIP and AlexNet\autocite{krizhevsky2012} (see \edfig{1} for visualizations of these principal components). We reasoned that CLIP would be a strong candidate for creating high-quality coarse categories given its language-aligned pretraining, which may organize its representations along semantically informative axes. However, we observed similar general trends when training on coarse categories derived from other source networks, including AlexNet, as shown in the main figures, and ViT\autocite{dosovitskiy2021} and DINOv3, as shown in \edfig{2}.

Additionally, although our main analyses rely on coarse categories defined from pretrained models, our key findings are not contingent on the use of pretrained networks. First, we observed qualitatively similar results when using handcrafted coarse category labels based on binary distinctions along interpretable semantic dimensions (Fig.~\ref{fig:semantic}; Section~\ref{sec:semantic}). Second, we observed similar trends when defining coarse categories based on superordinate labels from the WordNet hierarchy (\edfig{3}). However, unlike semantically defined categories or the WordNet hierarchy, our data-driven procedure is highly generalizable and can define categories from any pretrained embeddings in any modality. Here, we leveraged this data-driven approach to derive coarse categories from ImageNet-trained models as well as models trained with vision--language alignment and self-supervised objectives.

% Additionally, although we derived coarse category labels using a data-driven approach, our findings do not depend on privileged access to a pretrained network: eight classes defined by hand from three semantic dimensions closely approach the alignment of the data-driven eight-class models (Fig.~\ref{fig:semantic}; Section~\ref{sec:semantic}), and networks trained on superordinate categories from the WordNet noun hierarchy show qualitatively similar results (\edfig{3}). However, unlike the fixed WordNet hierarchy, our data-driven procedure is highly generalizable and can define categories from any pretrained embeddings in any modality. Here we applied this procedure to ImageNet-trained models as well as models trained with vision--language alignment and self-supervised objectives.

\section{Measuring biological alignment}\label{sec:measuring-alignment}

We compared network representations with biological representations obtained from three sources: human fMRI, macaque electrophysiology and human similarity judgments. Because the units of these systems (voxels, electrodes, behavioral embedding dimensions, and network features) do not correspond to one another, we used representational similarity analysis (RSA) to compare systems at the level of representational geometry\autocite{kriegeskorte2008}. For each system, we computed a representational dissimilarity matrix (RDM) whose entries are the correlation distances (one minus the Pearson correlation) between the response patterns evoked by each pair of stimuli. Alignment between a network layer and a brain region or behavioral embedding was quantified as the Spearman rank correlation between the upper triangles of their RDMs. This measure asks whether two systems agree on which stimuli are relatively similar and which are distinct, without fitting any parameters to the biological data (\methods{sec:rsa}{Representational similarity analysis}).

Each network has many candidate layers. To avoid selecting layers on the same data used for reporting, we split every evaluation dataset into a selection partition and a reporting partition of disjoint stimuli (for the behavioral data, disjoint concepts). For each region and model, we identified the layer with the highest mean RSA across participants in the selection partition; the final RSA score was obtained by examining this selected layer in the reporting partition (\methods{sec:layer-selection}{Layer selection on independent stimuli}). Neural responses were restricted to reliable units, retaining fMRI voxels with a noise-ceiling signal-to-noise ratio above 0.2 and recording electrodes with an across-repetition reliability above 0.3. Confidence intervals were estimated by resampling reporting stimuli with replacement (1,000 iterations), with participants and training seeds held fixed. To test whether the results depend on the unweighted geometry, we also fitted ridge-regression encoding models that linearly reweight network features to predict individual voxel responses (\edfig{6}).

\section{Coarse-grained training produces a more categorical representational geometry}\label{sec:different-representations}

We first asked how coarse supervision changes the internal geometry of the representations a network learns. To do this, we compared two networks: the pretrained AlexNet whose principal components defined a set of four coarse category labels, and an AlexNet-style network trained from scratch on those labels. For each network, we extracted activations from the first fully connected layer for 50,000 ImageNet images and projected them onto that network's first two principal components (Fig.~\ref{fig:labels}b,c).

In the pretrained network, the four coarse classes overlap. They were defined from this network's representation, so they are present in it, but they are not what organizes it. In the coarse-trained network, the four classes are pulled apart and occupy separate regions of the projection. This is expected: the network is trained to tell these four groups of images apart, so it learns a representation in which they are well separated, resulting in a more categorical geometry. The point of the comparison is that the coarse-trained representation is not a smoothed or lower-dimensional version of the one learned under 1,000-way classification. Training on coarse labels changes how the representation is organized, not just how finely it is divided. We test this directly later, by asking whether reducing the 1,000-class representation to its leading principal components reproduces the alignment of coarse-trained networks (\edfig{7}).

\begin{figure}
\centering
\includegraphics[width=\linewidth,keepaspectratio]{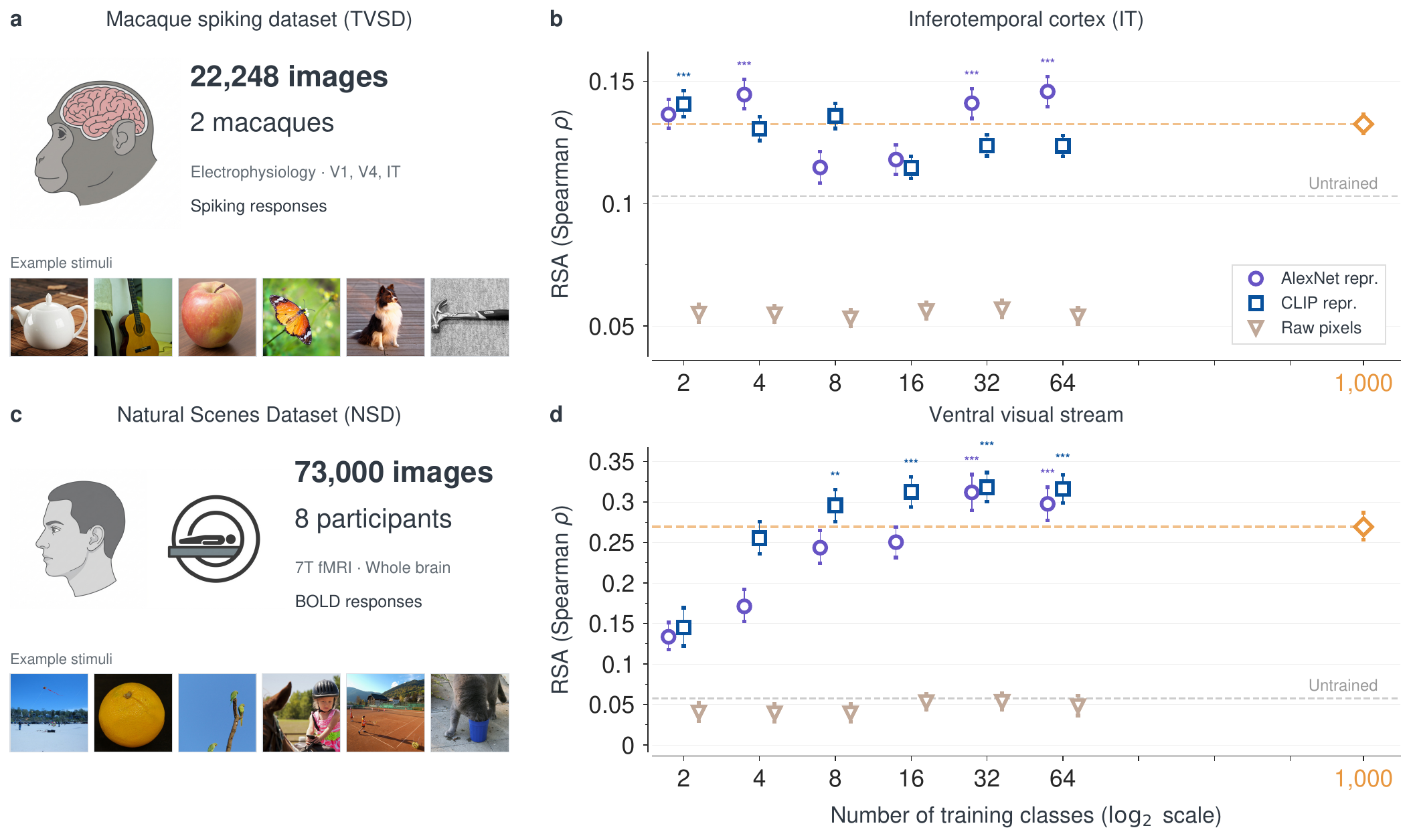}
\caption{\textbf{Coarse supervision is sufficient to match or exceed the cortical alignment achieved by 1,000-class training.}
\textbf{a,c}, Neural datasets used to evaluate network representations: responses to object images recorded from two macaques (TVSD; \textbf{a}) and fMRI responses to natural scenes from eight human participants (NSD; \textbf{c}).
\textbf{b,d}, Representational alignment with macaque inferotemporal cortex (\textbf{b}) and the human ventral visual stream (\textbf{d}) for AlexNet-style networks trained on 2--64 coarse classes. Coarse labels are derived from principal components of pretrained AlexNet representations (purple circles), pretrained CLIP representations (blue squares) or raw pixels (tan triangles). All networks are trained from scratch on ImageNet. The orange diamond and dashed line indicate the same architecture trained on the original 1,000 categories; the grey dashed line indicates an untrained network. Alignment is measured using representational similarity analysis (RSA): the Spearman correlation between network and neural dissimilarity matrices, with higher values indicating more similar organization. Each network's best-aligned layer is selected using separate images. Training on a small number of representation-derived classes can match or exceed 1,000-class alignment in both species, whereas pixel-derived classes remain below the untrained baseline.
Points are means across individuals and three training seeds; error bars are 95\% confidence intervals from 1,000 stimulus-bootstrap resamples. Asterisks indicate significantly higher alignment than the 1,000-class baseline in a two-sided paired stimulus bootstrap, after correction for multiple comparisons (false discovery rate, $q < 0.05$). Statistical details are provided in \methods{sec:paired-comparisons}{Paired comparisons with the 1,000 class baseline}.}\label{fig:neural}
\end{figure}
\section{A coarse supervisory signal is sufficient for high neural alignment}\label{sec:neural-alignment}

Given that coarse and fine-grained training produce qualitatively different representational geometries, we next asked how these representations compare with biological vision. Because we trained all networks from scratch with identical architectures, datasets, and hyperparameters, varying only the number of training classes, we could isolate its effect without the many confounds that typically complicate comparisons across models (e.g., augmentation schemes, learning rates, normalization; \methods{sec:architectures-training}{Network architectures and training}). Throughout, the fine-grained reference is a 1,000-class baseline trained with the same architecture, training recipe, image corpus and train/test partition as the coarse networks, differing only in its labels, which are the original ImageNet categories. All networks were trained with three random seeds to ensure that effects are not contingent on initialization conditions.

We evaluated neural alignment against two complementary datasets spanning different scales of neural measurement (Fig.~\ref{fig:neural}). The Natural Scenes Dataset (NSD) contains 7T fMRI responses from eight human subjects viewing $\sim$10,000 natural images\autocite{allen2022}. The THINGS Ventral-stream Spiking Dataset (TVSD) contains multi-unit spiking responses from macaque visual cortex to $>$22,000 object images, offering a high-resolution measurement of cortical responses in a region critical for object recognition\autocite{papale2025}. In both datasets, we focus on image-evoked responses from high-level regions along the ventral visual stream; early visual regions are reported in \edfig{8} (\methods{sec:datasets}{Neural and behavioral datasets}).

We used RSA to compare the representations in these regions with neural networks trained to distinguish different numbers of classes. Typical ImageNet supervision involves training a network to discriminate among 1,000 fine-grained categories. We sought to determine how many coarse-grained categories networks need to be trained on to reach the same level of brain alignment as a 1,000-way trained network. Our findings show that a remarkably coarse-grained supervised signal suffices. Specifically, in both the human and monkey data, we found that networks trained on a small number of broad categories, as few as eight, matched or surpassed the brain alignment of a 1,000-way supervised network. This holds in high-level visual cortex in both species (Fig.~\ref{fig:neural}). The significance markers in Fig.~\ref{fig:neural} come from a paired stimulus bootstrap; for the human data, a complementary paired $t$-test across the eight participants identifies the same set of coarse networks as significantly exceeding the baseline (\methods{sec:paired-comparisons}{Paired comparisons with the 1,000 class baseline}). In early visual cortex, by contrast, alignment depends little on training: trained networks lie close to the untrained network in the human early visual stream, and in macaque V1 the untrained network is the best-aligned model (\edfig{8}). This is consistent with prior reports that randomly initialized convolutional networks match or exceed trained networks in V1 and early visual cortex while falling far behind them in higher areas\autocite{cichy2016,farahat2025,kazemian2025,miao2024convolutional}, reflecting the strong architectural prior that convolution and pooling impose on early visual features. The findings are consistent across networks initialized with different random seeds, as shown for the human fMRI and behavioral data in \edfig{4}, and they hold when the coarse-scale category labels are derived from other supervised and self-supervised networks (\edfig{2}), indicating that the effect does not depend on the particular embedding of any single source model.

\edfig{5} shows the accuracy of the AlexNet-style networks on each network's own $K$-class task, measured on a held-out partition of the training corpus, for each label source; accuracies for the larger architectures are listed in \methods{sec:training-outcomes}{Training outcomes}. Although classification accuracy varies across levels of coarseness and across different source networks, these differences in classification accuracy do not resemble the differences in neural alignment for these same networks, suggesting that the alignment results cannot be simply explained by classification performance\autocite{linsley2023,storrs2021}. Furthermore, we observed similar trends in the neural alignment results when comparing networks to neural data using encoding models instead of RSA, thus demonstrating that even with linear reweighting, coarse-supervision suffices to reach the performance of 1,000-way training (\edfig{6}; \methods{sec:encoding}{Encoding models}).

The visualization analyses in Fig.~\ref{fig:labels} suggest that coarse-supervised networks learn representations with a qualitatively different geometry than their corresponding source networks. This suggests that the brain alignment of coarse-supervised networks cannot be reproduced through a simple dimensionality reduction of a source network's representations. To test this, we reconstructed representations from both coarse- and fine-grained models using only the top-k principal components and measured neural alignment as a function of k. If coarse-grained representations occupied a low-dimensional subspace of the fine-grained representational space, alignment scores should converge at low k. They do not (\edfig{7}), confirming that coarse-grained training does not merely discard dimensions from a richer representation but gives rise to a distinct representational geometry.

Additional follow-up experiments demonstrate that not any coarse partitioning suffices. When we carved up the input space using raw pixel statistics, producing categories based on low-level image features such as luminance and contrast, the resulting representations showed weak neural alignment (Fig.~\ref{fig:neural}; \methods{sec:pixel-control}{Pixel PCA control}). 

\begin{figure}
\centering
\includegraphics[width=\linewidth,keepaspectratio]{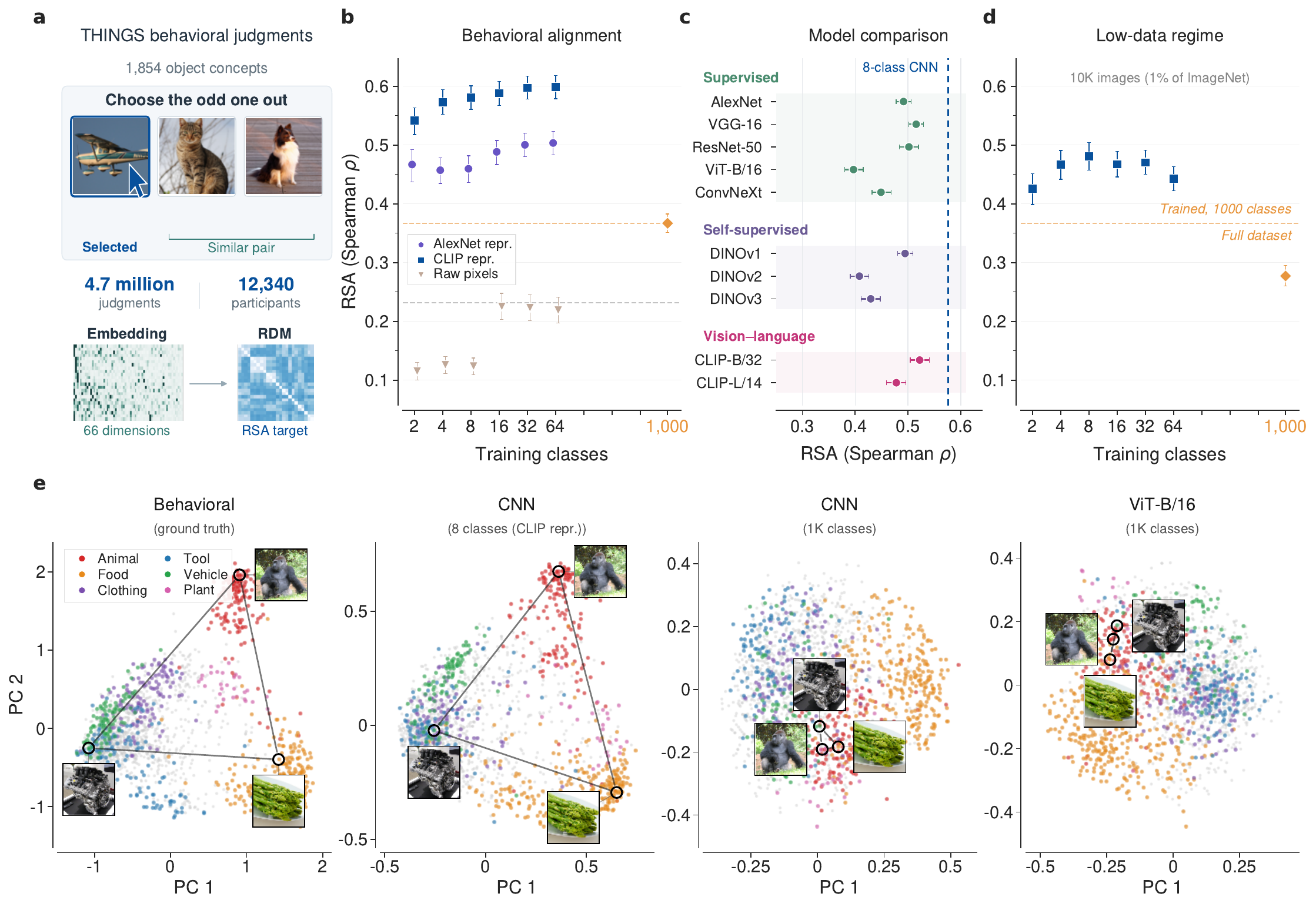}
\caption{\textbf{Coarse supervision signals produce stronger alignment with human visual similarity judgments than all other models.}
\textbf{a}, The THINGS behavioral benchmark. Participants choose the odd one out from three object images, indicating which pair is most similar. A 66-dimensional embedding fitted to 4.7 million judgments from 12,340 participants describes similarities among 1,854 object concepts and defines the behavioral representational dissimilarity matrix (RDM).
\textbf{b--d}, Behavioral alignment is the Spearman correlation between model and behavioral RDMs; higher values indicate closer agreement with human judgments. Model layers are selected using 370 concepts and evaluated on the remaining 1,484.
\textbf{b}, Alignment of AlexNet-style networks trained from scratch on ImageNet using 2--64 coarse classes derived from pretrained AlexNet representations (purple circles), pretrained CLIP representations (blue squares) or raw pixels (tan triangles), as in Fig.~\ref{fig:neural}. The orange diamond and dashed line indicate 1,000-class training; the grey dashed line indicates an untrained network.
\textbf{c}, Comparison with pretrained models using supervised classification, self-supervised learning (DINO) or vision--language learning (CLIP). The network trained on just eight CLIP-derived classes (vertical dashed line) achieves higher behavioral alignment than every pretrained model shown.
\textbf{d}, Training with limited data. Networks trained on 10,000 ImageNet images (approximately 1\% of the training set) using CLIP-derived coarse classes (blue squares) are compared with 1,000-class networks trained on the same images (orange diamond) or the full dataset (orange dashed line).
\textbf{e}, The same 1,854 concepts projected onto the first two principal components of the behavioral embedding, the first fully connected layer (FC1) of the eight-class and 1,000-class networks, and the penultimate layer of a supervised ViT-B/16, respectively. Colors identify six THINGS super-categories; other concepts are grey. Open circles and thumbnails highlight gorilla, engine and asparagus in each projection.
Error bars indicate 95\% confidence intervals from 1,000 bootstrap resamples of the held-out concepts.}\label{fig:behaviour}
\end{figure}
\section{Coarse supervision produces the most behaviorally aligned model}\label{sec:behavior}

Neural alignment captures how well a network's representations match the structure of brain activity, but does it also match how humans actually perceive and organize the visual world? To address this question, we examined network alignment with large-scale perceived similarity judgments from human observers. We used human similarity judgments from the THINGS dataset\autocite{hebart2019} (Fig.~\ref{fig:behaviour}). In these data, the perceived similarities among objects from different categories were assessed using a triplet odd-one-out paradigm: participants viewed three images and selected which was the odd one among the set. The task requires no verbal labels or predefined similarity dimensions. The dataset contains responses to $\sim$4.7 million odd-one-out trials from $\sim$12,000 participants, which were used to learn a set of 66-dimensional embeddings for 1,854 object concepts\autocite{hebart2023}. The scale and richness of this dataset make it ideally suited for evaluating whether networks organize visual information the way people do.

We compared neural network representations with these behavioral embeddings using RSA. Our findings revealed an unexpected result. Networks trained on coarse supervisory signals do not simply match the performance of a model trained on 1,000-way classification; they substantially exceed it (Fig.~\ref{fig:behaviour}b). This advantage is already present with as few as eight coarse classes. To contextualize this benefit of coarse supervision, we benchmarked against a broad set of pretrained models spanning diverse architectures (CNNs, Transformers) and training paradigms (supervised, self-supervised, large-scale pretrained), many of which were trained on far larger datasets than our models (Fig.~\ref{fig:behaviour}c). Across every network in this set, our coarse-supervised models achieved the highest alignment with human perceptual similarity. This suggests that coarse supervision is not merely sufficient but actually advantageous for capturing the high-level structure of human object perception.

To better understand why coarse supervision yields more human-aligned representations, we performed exploratory visualization analyses by plotting principal component projections of behavior-derived and model representations for all 1,854 object concepts from the THINGS dataset (Fig.~\ref{fig:behaviour}e). These plots reveal a salient categorical structure present in both the behavior-derived and coarse-supervised representations but absent from conventional networks trained on 1,000-way classification\autocite{jozwik2017}. Both humans and coarse-supervised networks exhibit strong distinctions among broad high-level categories such as animals, food, and vehicles, whereas conventional networks exhibit a more diffuse similarity space. Notably, the coarse-supervised models exhibit these high-level semantic distinctions without being explicitly trained on the category labels plotted here, or even on these images. Together, these analyses suggest that the improved behavioral alignment of coarse-supervised networks reflects their emergent clustering of images into broad, behaviorally relevant domains.

\section{Coarse supervision achieves high alignment even with limited data}\label{sec:limited-data}

The preceding results establish that coarse supervision produces higher behavioral alignment than fine-grained training when both have access to the full dataset. But how much data does a coarsely trained model actually need? If coarse categories capture the broad perceptual distinctions that matter most, the learning problem may be inherently simpler, requiring fewer examples to arrive at a human-aligned representation.

To test this, we trained models from scratch using approximately 1\% of the ImageNet dataset, again comparing coarse and fine-grained supervisory signals with all other parameters held constant. Networks trained with a coarse supervisory signal on this small fraction of the data achieved higher behavioral alignment than a 1,000-class model trained on the full training set of roughly one million images (Fig.~\ref{fig:behaviour}d). Thus, the coarse-supervised networks exceeded the full-data 1,000-class baseline using approximately 1\% of its training images.

\begin{figure}
\centering
\includegraphics[width=\linewidth,keepaspectratio]{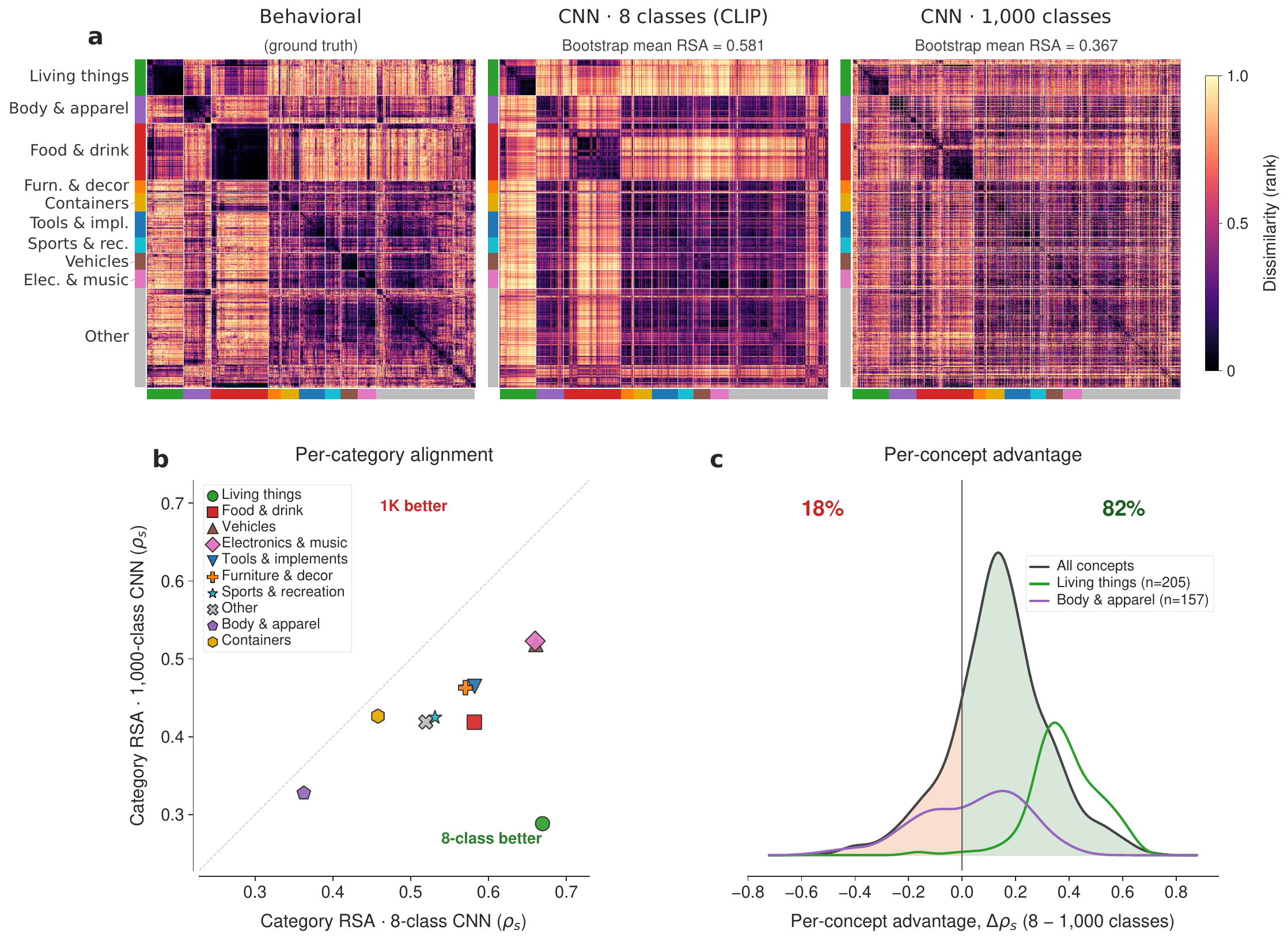}
\caption{\textbf{The advantage of coarse supervision holds across semantic categories and for the large majority of individual concepts.}
\textbf{a}, Representational dissimilarity matrices (RDMs) for all 1,854 THINGS concepts, comparing the human behavioral embedding (left), an AlexNet-style network trained on eight CLIP-derived coarse classes (middle) and a network trained on the 1,000 ImageNet categories (right). Dissimilarities are ranked and scaled from 0 to 1; darker colors indicate greater similarity. Concepts follow the same order in all three matrices, grouped into ten super-categories (colored bars, with remaining concepts grouped as `Other') and ordered within each group by hierarchical clustering of the behavioral RDM. Values above the network matrices report bootstrap mean RSA with the behavioral RDM.
\textbf{b}, Alignment within each super-category for the eight-class network (horizontal axis) and the 1,000-class network (vertical axis). Alignment is calculated for each concept as the Spearman correlation between corresponding rows of the network and behavioral RDMs, then averaged within each super-category. All ten groups fall below the line of equal alignment, indicating closer agreement with human judgments for the eight-class network.
\textbf{c}, Distributions of the difference in alignment for individual concepts ($\Delta\rho$ = eight-class $-$ 1,000-class), shown as kernel density estimates. Positive values favor the eight-class network, which has higher alignment for 82\% of concepts; negative values favor the 1,000-class network (18\%). Curves show all concepts (black), living things (green, $n$ = 205) and body and apparel (purple, $n$ = 157), the groups with the largest and smallest average improvements in \textbf{b}, respectively.}\label{fig:concepts}
\end{figure}
\section{Coarse training improves alignment across all semantic categories}\label{sec:semantic-categories}

The aggregate behavioral alignment scores show that coarsely trained models outperform fine-grained models overall, but is this driven by a few categories where coarse distinctions happen to suffice, or does it reflect a broad advantage across the full diversity of visual concepts?

To answer this, we decomposed alignment at the level of individual concepts and semantic categories. For each of the 1,854 object concepts in the THINGS dataset, we extracted its corresponding row from the representational dissimilarity matrix and computed how well that concept's similarity structure aligned with human judgments, yielding a per-concept alignment score for each model. We then compared the coarsely trained model against the 1,000-class model by taking the per-concept difference in alignment, revealing that over 80\% of the concepts are better captured by the coarse model (Fig.~\ref{fig:concepts}; \methods{sec:per-concept}{Per concept alignment}).

These concepts in the THINGS database include labels for higher-level semantic categories (e.g., animals, food, tools, vehicles), allowing us to average across constituent concepts to obtain category-level alignment scores. Across every semantic category, the coarsely trained model achieves higher alignment with human perceptual similarity than the fine-grained model (Fig.~\ref{fig:concepts}). This pattern can also be observed in the accompanying representational dissimilarity matrices for human behavior, the coarse model (8 classes), and the fine-grained model: the coarse model's similarity structure mirrors the block-diagonal organization of human perception across the full range of object categories.

\begin{figure}
\centering
\includegraphics[width=\linewidth,keepaspectratio]{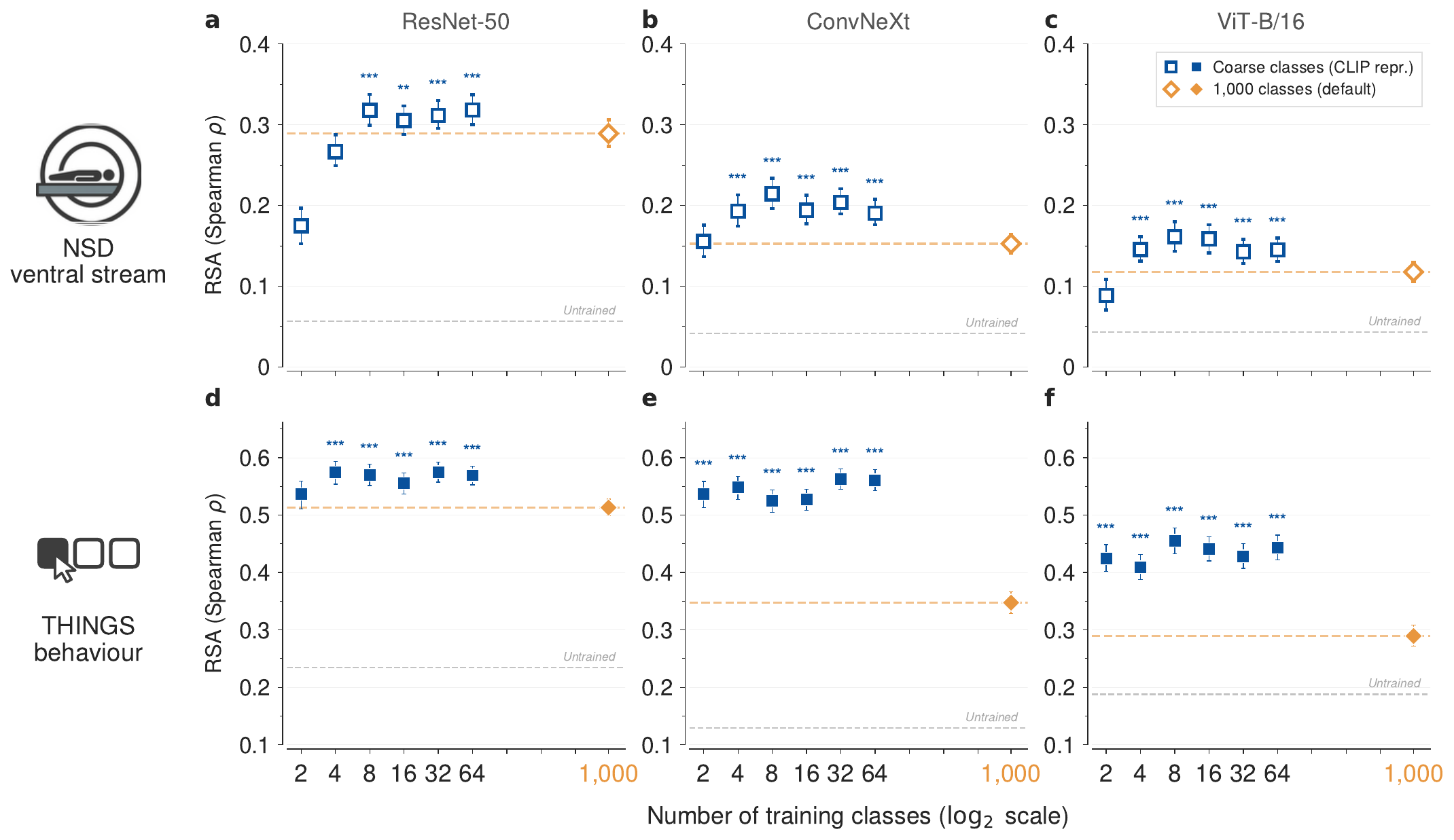}
\caption{\textbf{Coarse supervision improves alignment with human visual cortex and behavior across network architectures.}
\textbf{a}--\textbf{f}, Alignment for ResNet-50 (\textbf{a},\textbf{d}), ConvNeXt (\textbf{b},\textbf{e}) and ViT-B/16 (\textbf{c},\textbf{f}) trained from scratch on ImageNet. Blue squares indicate training on 2--64 CLIP-derived coarse classes; orange diamonds and dashed lines indicate the same architecture trained on the original 1,000 categories; grey dashed lines indicate untrained networks.
\textbf{a}--\textbf{c}, Alignment with human ventral visual stream responses (NSD; open symbols).
\textbf{d}--\textbf{f}, Alignment with human similarity judgments (THINGS; filled symbols). Alignment is measured by representational similarity analysis (RSA), the Spearman correlation between model and neural or behavioral dissimilarity matrices; higher values indicate closer agreement. Layer selection and evaluation follow Figs.~\ref{fig:neural} and \ref{fig:behaviour}. Each condition uses one trained network; NSD scores are averaged across eight participants. Error bars indicate 95\% bootstrap confidence intervals. Asterisks mark higher alignment than the corresponding 1,000-class network after correction for multiple comparisons ($q < 0.05$), with increasing numbers indicating stronger statistical evidence; test details and thresholds are provided in \methods{sec:paired-comparisons}{Paired comparisons with the 1,000 class baseline}.}\label{fig:architectures}
\end{figure}
\section{Coarse training generalizes across modern architectures}\label{sec:architectures}

All experiments presented thus far use a relatively simple CNN architecture with five convolutional layers and two fully connected layers followed by a linear classifier, similar to AlexNet. This raises a natural question: does the advantage of coarse supervision depend on the simplicity of the architecture, or does it persist in larger, more expressive models? To test this, we trained ResNet-50\autocite{he2016}, ConvNeXt\autocite{liu2022}, and a Vision Transformer (ViT) from scratch on ImageNet across the full range of coarse-to-fine supervisory signals. Within each architecture, every model was trained with one fixed recipe, so that only the number of training classes varies (\methods{sec:additional-architectures}{Additional architectures}). Across all three modern architectures, the pattern holds on both benchmarks (Fig.~\ref{fig:architectures}): coarse supervision produces representations with surprisingly high alignment to the human ventral visual stream (Fig.~\ref{fig:architectures}a--c) and to human perceptual judgments (Fig.~\ref{fig:architectures}d--f), matching or exceeding that of their 1,000-class counterparts (as in Fig.~\ref{fig:neural}, the participant-level test agrees with the stimulus bootstrap for the fMRI data). Thus, the advantage of coarse supervision is not an artifact of limited model capacity; it emerges consistently regardless of architectural family, model size, or hyperparameters.

\begin{figure}
\centering
\includegraphics[width=\linewidth,keepaspectratio]{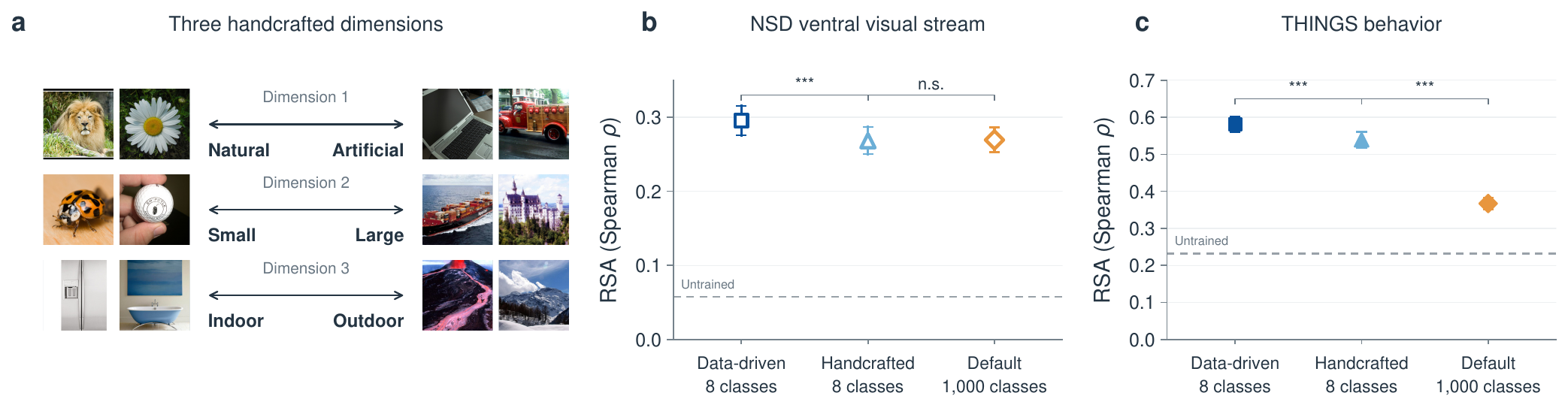}
\caption{\textbf{Eight handcrafted semantic classes are sufficient to learn representations aligned with human visual cortex and behavior.}
\textbf{a}, Eight classes defined by combining three semantic distinctions: artificial or natural, small or large real-world size (whether an object fits in the hands), and outdoor or indoor. Each ImageNet category is assigned to one combination, illustrated by an example image (\methods{sec:semantic-labels}{Explicit semantic labels}).
\textbf{b},\textbf{c}, Alignment of AlexNet-style networks with human ventral visual stream responses (\textbf{b}) and THINGS similarity judgments (\textbf{c}). Networks are trained from scratch on eight classes derived from pretrained CLIP representations (blue squares) or from the handcrafted distinctions in \textbf{a} (light-blue triangles). Orange diamonds indicate training on the original 1,000 categories; the grey dashed line indicates an untrained network. Alignment is measured by representational similarity analysis (RSA; Spearman $\rho$); higher values indicate closer agreement with human representations. Handcrafted classes yield similar cortical alignment and higher behavioral alignment than 1,000-class training. Points show means across three training seeds and, in \textbf{b}, eight participants. Error bars indicate 95\% bootstrap confidence intervals. Brackets compare handcrafted classes with the indicated alternatives after correction for multiple comparisons ($q < 0.05$): ***, $P < 0.0025$; n.s., not significant.}\label{fig:semantic}
\end{figure}
\section{Handcrafted semantic categories suffice as a coarse supervisory signal}\label{sec:semantic}

So far, the coarse labels have come from a data-driven procedure. We next asked whether a set of coarse categories could instead be specified from known organizing principles of human vision. As a test case, we built eight categories from three binary distinctions that are among the best-documented organizing dimensions of the ventral visual stream: natural versus artificial\autocite{kriegeskorte2008,konkle2013}, small versus large real-world size\autocite{konkle2012,konkle2013} and indoor versus outdoor\autocite{henderson2007} (Fig.~\ref{fig:semantic}a). Each of the 1,000 ImageNet classes was assigned to one of the eight categories on the basis of its class name, first by a large language model and then checked by manual review, so every image in a class receives the same coarse label (\methods{sec:semantic-labels}{Explicit semantic labels}). The label carries three bits, the same as the eight-class data-driven condition, but with the labels derived from semantic categorization rather than PCA of pretrained embeddings. 

A network trained on these eight handcrafted classes matched the alignment of 1,000-way training when comparing with the human ventral visual stream and exceeded it when comparing with human similarity judgments, though interestingly, its performance was slightly short of the network trained on eight CLIP-derived classes on both the neural and behavioral benchmarks (Fig.~\ref{fig:semantic}b,c). This proof of principle finding shows that the effectiveness of coarse supervision does not depend on the use of pretrained embeddings, as coarse supervision is effective even when the categories are derived from intuitive semantic labels. Nonetheless, the data-driven procedure remains our main approach because it requires no annotation and lets us vary the number of classes parametrically, allowing us to rigorously examine the effect of coarseness on the learned representations.

\section{Discussion}\label{discussion}

While the field has largely focused on self-supervised learning\autocite{konkle2022,margalit2024,zhuang2021} and fine-grained classification\autocite{yamins2014,yamins2016,kubilius2019} as the dominant paradigms for producing brain-aligned representations, we have shown that networks trained on just a handful of broad categories match or surpass fine-grained supervised models in alignment with the visual cortex representations of both humans and monkeys. Furthermore, in the case of behavioral alignment, our findings reveal that coarse-grained supervision is not only sufficient but notably advantageous: coarse-supervised models more closely match human intuitions about perceptual similarity than all networks tested, including leading models trained on much larger datasets. Together, these findings reveal that learning to cluster natural images into extremely coarse categories is sufficient for networks to develop internal representations that mirror human vision.

Our findings provide a new perspective on the question of what kinds of optimization objectives might shape biological vision\autocite{doerig2025,yamins2016,doerig2023,dwivedi2021,conwell2024,gokce2024,xie2024,wang2019,chen2025}. Previous work has proposed that neural networks converge to brain-aligned representations when they are optimized on challenging tasks that parallel the specific behavioral goals that biological vision has evolved to support (e.g., invariant object recognition)\autocite{yamins2016,doerig2023,cao2024}. Our findings show that the optimization objectives needed to learn brain-aligned representations may be far more rudimentary than previous work suggests. Instead of optimizing on the fine-grained recognition of specific object categories or even finer-grain instance-level recognition, we find that a much coarser objective involving as few as eight broad domains of images can yield similar or even better human alignment. Thus, our results suggest that a promising direction for future work is to uncover the simplest forms of optimization that suffice for converging to human-like representations\autocite{passi2026}, and to examine whether such coarse learning objectives might serve to establish a foundation model for vision, composed of general-purpose features that can support many downstream tasks\autocite{cusack2024}.

We used pretrained model representations to derive coarse category labels. This data-driven approach allowed us to define categories based on the inherent structure of natural image statistics and to systematically vary the number of training classes. It also has the benefit of being highly general and applicable to any pretrained model (including different supervised and self-supervised models and different data modalities). We do not claim that these principal components are the only coarse dimensions that work, or the best ones. The handcrafted labels in Fig.~\ref{fig:semantic} show, as a proof of principle, that a different set of coarse distinctions also produces human-aligned representations, though the data-driven labels remain more effective and require no annotation. An open question is where such broad distinctions might come from in biological vision. One possibility is to use behavioral experiments to reveal potential hierarchical structure in human image representations (either explicitly through image annotations or implicitly through similarity judgments). Another possibility is to take inspiration from developmental research and to derive broad categories based on the kinds of distinctions that the visual system makes in infancy\autocite{spriet2022,spelke2007,ayzenberg2025}.

We showed that neural networks can learn a deep hierarchy of human-aligned visual features through coarse supervision. However, human visual representations ultimately support fine-grained recognition across many more categories, and this capacity clearly requires learning beyond what a handful of superordinate distinctions can provide. Thus, a direction for future work is to understand how coarse and fine-grained learning interact\autocite{chen2018,xu2021,zhang2023}. One possibility is a hierarchical curriculum in which coarse supervision builds the representational scaffold and finer-grained learning subsequently elaborates within-category structure on top of it\autocite{stretcu2021}. Prior work using progressive training along the WordNet hierarchy shows promising results in this direction\autocite{ahn2021,peterson2018,su2021,wang2016,pirovano2025}. Our coarse-graining method provides a new data-driven framework for constructing such curricula based on the natural statistics of sensory stimuli. Exploring whether such data-driven curricula might outperform the WordNet hierarchy is a promising next step. Furthermore, one could adapt this approach to other data modalities, such as auditory networks or even language models (e.g., instead of predicting the next word, a model could be trained to predict the next coarse category).

Our findings also suggest a new direction for building AI systems with more human-aligned perception. Recent work has sought to improve the human alignment of neural networks by directly optimizing on large-scale human behavioral data\autocite{muttenthaler2023,muttenthaler2025,fu2023,fel2022}. Our coarse-supervised models improved alignment with human similarity judgments without training on those judgments. Given that coarsening the supervisory signal requires no manual annotation, our framework can be readily scaled up and applied to even larger models and training sets, and it can be adapted to any pretrained model in any modality. Thus, our approach opens new possibilities for exploring whether coarse-scale supervision might improve human alignment across a wide range of AI models.

Our findings raise new questions about whether there may be biological mechanisms for delivering such a coarse supervisory signal in both the mature and developing brain. There is evidence that several feedback pathways to visual cortex may carry low-dimensional categorical information: prefrontal neurons encode category membership and project to inferotemporal cortex\autocite{freedman2001}, dopaminergic reward signals selectively modulate stimulus representations in visual cortex\autocite{arsenault2013}, and the amygdala sends topographically organized feedback projections to all levels of the ventral visual stream\autocite{amaral2003}. The thalamus is a promising candidate as a locus of coarse feedback signals in development, and its connections with visual cortex already exhibit topographic organization in neonates\autocite{ayzenberg2025}. Furthermore, developmental neuroimaging has shown that several high-level visual categories are represented in visual cortex as early as two months of age\autocite{odoherty2026}. This suggests that an exciting direction for future work is to test whether the developing brain might contain innate inductive biases for carving the visual world along its natural axes to learn broad categorical distinctions.

\section{Data and code availability}\label{data-code-availability}

All datasets analyzed in this study are publicly available and were obtained from their original sources: ImageNet\autocite{deng2009}, the Natural Scenes Dataset\autocite{allen2022}, the macaque ventral-stream spiking dataset\autocite{papale2025}, and the THINGS object images\autocite{hebart2019} together with the THINGS-data behavioral similarity judgments\autocite{hebart2023}. No new data were collected for this work. All code used to derive the coarse labels, train the networks and run every alignment analysis is available at \url{https://github.com/yashsmehta/visreps}, along with the derived label assignments. The database of evaluation results underlying every figure, and every network trained from scratch here (the AlexNet-style CNNs and the ResNet-50, ConvNeXt and ViT-B/16 models, at each label granularity and seed) will be made available to download.

\printbibliography

% Methods and Extended Data follow the references so that a single main.tex builds the whole document.
\clearpage
% Methods. Body only: included by main.tex after the references.
% Headings are unnumbered; hyperref still anchors each \label, so \methods{label}{title} links resolve.
\setcounter{secnumdepth}{0}
\section{\Large Methods}\label{sec:methods}

\section{Overview and terminology}\label{sec:overview}

We asked how the number of training classes shapes the representations learned by visual networks. Randomly initialized networks were trained to predict either the original 1,000 ImageNet categories or a smaller set of categories defined by a fixed mapping from images to labels. The primary label family was constructed by partitioning the representations of a pretrained network along its principal components. Additional label families used raw pixel principal components, WordNet ancestors, or explicitly annotated semantic attributes.

Two roles are distinguished throughout. A \textbf{source model} is a pretrained network whose representations are used only to assign labels to images. A \textbf{trained model} is a network trained from scratch to predict those labels. No source model weights are transferred to trained models.

For clarity, \textbf{classification training images} are the images used to optimize network weights. \textbf{Selection images} are a partition of each evaluation dataset used to choose network layers (and, for encoding, to fit readouts). \textbf{Reporting images} are the held out evaluation partition on which all scores are computed. Network weights are frozen during evaluation.

\section{Classification data and supervisory labels}\label{sec:labels}

\subsection{ImageNet corpus and split}\label{sec:imagenet-split}

Training used a local ImageNet corpus of 1,261,406 images in 1,000 categories, corresponding to the ILSVRC 2010 category set; category identities were taken from the corpus's WordNet identifiers. A fixed random permutation assigned 80\% of images (1,009,124) to classification training and 20\% (252,282) to classification evaluation. This is an image level random split of the local corpus, not the official ILSVRC 2012 training/validation split. Coarse labels were defined using the complete corpus, including the PCA basis and median thresholds, whereas weight updates used only the training partition.

\subsection{Source representations}\label{sec:source-representations}

Source features were extracted with each source model in evaluation mode using that model's standard preprocessing.

\begin{table}[H]
\centering
\begin{tabular}{lllr}
\toprule
Source & Implementation & Representation & Dimension \\
\midrule
AlexNet & torchvision, ImageNet weights & Second hidden fully connected layer, post ReLU & 4,096 \\
CLIP & OpenAI ViT-L/14 & Projected image encoder output & 768 \\
DINOv3 & ViT Large, patch 16 & Final class token & 1,024 \\
Supervised ViT & ViT Large, patch 16, 224 px & Final class token & 1,024 \\
\bottomrule
\end{tabular}
\end{table}

Each feature vector was normalized to unit Euclidean norm before PCA, \(x_i = f_i / \|f_i\|_2\), so that partitions depend on the direction of the source representation rather than its magnitude. This normalization applies only to label construction; trained model activations were not renormalized during evaluation.

\subsection{Principal component partitions}\label{sec:pc-partitions}

Let \(X \in \mathbb{R}^{N \times D}\) hold the normalized source features. The covariance \(C = \frac{1}{N-1}\sum_i (x_i-\mu)(x_i-\mu)^\top\) was accumulated over the full corpus in double precision without featurewise standardization or whitening, and its eigenvectors \(v_j\) were ordered by decreasing eigenvalue. Each image received a score \(z_{ij} = (x_i-\mu)^\top v_j\) on axis \(j\), and a bit \(b_{ij} = \mathbf{1}[z_{ij} > m_j]\), where \(m_j\) is the median of \(z_{ij}\) across all images (ties receive zero). The \(K = 2^n\) class label combines the first \(n\) bits,

\[
y_i^{(n)} = \sum_{j=1}^{n} 2^{\,n-j}\, b_{ij},
\]

with PC1 as the most significant bit. Thus the 2 class condition uses PC1, the 4 class condition adds PC2, and so on. Axes and thresholds are global: PCA is fit once on the full corpus and medians are computed once per axis, not refit within previously defined groups. Removing the last bit recovers the coarser partition, so label systems are nested within a source model. The main sweep used \(K \in \{2, 4, 8, 16, 32, 64\}\).

Each individual bit is approximately balanced, but joint classes need not be, because orthogonal axes need not have independent median thresholded signs. \(K\) is the number of output classes; realized class occupancies are properties of each label set (CLIP-derived labels: 94,073 to 221,129 images per class at \(K=8\) and 2,040 to 64,403 at \(K=64\); AlexNet-derived: 123,254 to 195,694 and 4,623 to 41,144). We note that \(\log_2 K\) bounds the entropy of the label, not the knowledge embedded in the source model that shaped the partition.

\subsection{Pixel PCA control}\label{sec:pixel-control}

Each RGB image was resized to a 64 pixel shorter edge, center cropped to 64 $\times$ 64, scaled to \([0,1]\) without channel or norm normalization, and flattened to 12,288 dimensions. PCA and the global median rule above were applied to the leading six components. This condition tests partitions based on low resolution color and spatial structure with no learned representation.

\subsection{Explicit semantic labels}\label{sec:semantic-labels}

Each of the 1,000 ImageNet categories was annotated on three binary semantic attributes: natural versus manufactured, small versus large real-world size, and indoor versus outdoor. The annotations for each ImageNet category were applied to all images within that category. An initial set of annotations was generated using a large language model. For real-world size, we specified that anything small enough to be held in a human hand should be labeled "small." We then manually reviewed and edited these annotations to create the final set. 

% Each of the 1,000 ImageNet categories was annotated on three binary attributes, and the code was transferred to every image in the category: natural versus manufactured or prepared (prepared foods coded as artificial), handheld by an adult versus larger (a categorical judgment of the typical object), and typically indoor versus outdoor. The target is a single eight way label, \(y_i = 4\,\text{natural}_i + 2\,\text{handheld}_i + \text{indoor}_i\), trained as one classifier rather than three binary heads. Each category was first assigned on all three attributes by a large language model from its class name, and every assignment was then checked by manual review of category montages; CLIP ViT-L/14 indoor/outdoor scores were used to flag categories for reinspection, and this review changed indoor/outdoor labels for 66 categories and handheld labels for four. Revised classes contain 38,650 to 326,242 images (full corpus counts). Training used natural class frequencies without balancing.

\subsection{WordNet categories}\label{sec:wordnet-categories}

Each category's synset was mapped to its ancestor at a specified depth along a longest hypernym path to the root; synsets with shorter paths retained their deepest available node. Unique ancestors were assigned consecutive labels. Depths one through seven were generated; class counts at each depth are set by the taxonomy and need not be powers of two.

\section{Network architectures and training}\label{sec:architectures-training}

\subsection{Primary convolutional network}\label{sec:primary-cnn}

The primary model is an AlexNet style CNN with five convolutional blocks, adaptive average pooling, and two hidden fully connected layers.

\begin{table}[H]
\centering
\begin{tabular}{lrrrrrl}
\toprule
Block & In & Out & Kernel & Stride & Pad & Max pool after \\
\midrule
1 & 3 & 96 & 11 & 4 & 2 & 3 $\times$ 3, stride 2 \\
2 & 96 & 256 & 5 & 1 & 2 & 3 $\times$ 3, stride 2 \\
3 & 256 & 384 & 3 & 1 & 1 & none \\
4 & 384 & 384 & 3 & 1 & 1 & none \\
5 & 384 & 256 & 3 & 1 & 1 & 3 $\times$ 3, stride 2 \\
\bottomrule
\end{tabular}
\end{table}

Convolutions are ungrouped, bias free, and followed by BatchNorm and ReLU. The final feature map is pooled to 3 $\times$ 3 (2,304 features) and passed through two 4,096 unit layers, each with BatchNorm and ReLU and preceded by dropout (p = 0.3). A final linear layer outputs \(K\) logits. Convolutional and hidden weights use Kaiming normal initialization (fan out, ReLU); the classifier uses a zero mean normal with standard deviation \(1/\sqrt{D_{\text{in}}}\); biases are zero. Only the classifier size changes with the number of training classes.

\subsection{Optimization}\label{sec:optimization}

Primary networks were trained with AdamW (learning rate \(5\times10^{-4}\), weight decay \(10^{-3}\) applied to weight tensors only, batch size 32) for 20 epochs. The learning rate rose linearly from 25\% to 100\% of its nominal value over two warmup epochs, then followed cosine annealing to 5\% over the remaining 18 epochs, updated once per epoch. Training used mixed precision with gradient scaling and global norm clipping at 1.0. The loss was multiclass cross entropy with label smoothing \(\epsilon = 0.1\),

\[
q_{ik} = (1-\epsilon)\,\mathbf{1}[k = y_i] + \epsilon/K,\qquad
\mathcal{L} = -\frac{1}{B}\sum_{i}\sum_{k} q_{ik}\log p_{ik}.
\]

Augmentation was deliberately mild: resize to a 256 pixel shorter edge, center crop to 224 $\times$ 224, horizontal flip with probability 0.5, and rotation within $\pm$10$^\circ$, followed by ImageNet channel normalization. Classification evaluation used resize and center crop only. Preprocessing and the image split were identical across label conditions.

\subsection{Replicates and controls}\label{sec:replicates}

The main design crosses six class counts and five label sources with three training seeds (1, 2, 3), plus matched 1,000 class baselines. Epoch 20 checkpoints are the standard endpoint; epoch 0 checkpoints provide architecture matched untrained controls. Training set the PyTorch seed and used deterministic cuDNN settings.

\subsection{Training outcomes}\label{sec:training-outcomes}
Classification accuracy was measured on the 20\% held-out partition of the corpus after the final epoch, with no early stopping. The official ILSVRC 2012 validation set was not used: accuracy serves only as a check that each network learned its task, and because the coarse labels are defined on this corpus, a held-out slice of it is the only set on which every label condition and its chance level are defined. Every network learned its task far above chance, and accuracies agree across seeds within 2 percentage points (\edfig{5}). The 1,000 class networks reached the following top-1 accuracies on the held-out partition:

\begin{table}[H]
\centering
\begin{tabular}{lr}
\toprule
Architecture & 1,000 class top-1 accuracy \\
\midrule
Primary CNN (AlexNet style) & 74\% \\
ResNet-50 & 73\% \\
ConvNeXt-Base & 78\% \\
ViT-B/16 & 67\% \\
\bottomrule
\end{tabular}
\end{table}

The primary CNN value is the mean over three seeds; the other architectures were trained once, and their accuracies were recomputed from the evaluated epoch 20 checkpoints on the same partition.

\subsection{Additional architectures}\label{sec:additional-architectures}
ResNet-50, ConvNeXt-Base, and ViT-B/16 were trained from random initialization with their final classifiers sized to \(K\). All used batch size 32, label smoothing 0.1, mixed precision, and standard augmentation (RandomResizedCrop to 224 $\times$ 224, horizontal flip, ImageNet normalization). Within each architecture a single recipe was used for every number of classes, so that only \(K\) varies:
\begin{table}[H]
\centering
\footnotesize
\setlength{\tabcolsep}{3.5pt}
\begin{tabular}{llrrrrlr}
\toprule
Architecture & Optimizer & LR & Weight decay & Horizon & Warmup & Schedule & Clip \\
\midrule
ResNet-50 & SGD, momentum 0.9 & 0.015 & \(10^{-4}\) & 20 ep & 0 & Cosine & none \\
ConvNeXt-Base & AdamW & \(3\times10^{-4}\) & 0.05 & 20 ep & 2 & Cosine & none \\
ViT-B/16 & AdamW & \(3\times10^{-4}\) & 0.3 & 90 ep & 5 & Cosine & 1.0 \\
\bottomrule
\end{tabular}
\end{table}
All architectures were evaluated at the epoch 20 checkpoint; for ViT-B/16 this falls partway through its 90 epoch cosine schedule. One seed was trained per condition.

\subsection{Reduced data experiments}\label{sec:reduced-data}

Reduced data models were trained from scratch on a fixed subset of ten images per ImageNet category (10,000 images, about 1\% of the training partition) across the same seven label conditions (2 to 64 classes and 1,000 classes), single seed, for 100 epochs with the primary recipe otherwise unchanged, and evaluated with the same behavioral pipeline as the main models.

\section{Neural and behavioral datasets}\label{sec:datasets}

\subsection{Natural Scenes Dataset (NSD)}\label{sec:nsd}

We used the 1.8 mm release of NSD (eight participants, 7 T fMRI, natural scene photographs) and the released GLMsingle single trial betas (fitted HRF, GLMdenoise, ridge regularized estimates). Betas were z scored per voxel across presentations within each session and averaged over the available repetitions of each image. Regions were the early and ventral visual streams from the NSD streams parcellation. Within each ROI, voxels were retained only if their participant specific noise ceiling SNR exceeded 0.2, a reliability criterion independent of any model.

The shared images seen by all participants form the reporting set; each participant's remaining images form that participant's selection (and encoding training) set. Multi participant RSA uses the intersection of shared images available across participants, giving 907 reporting images with all eight participants. Encoding fits are participant specific and use that participant's full training set (8,302 to 9,000 images) and shared set. After the reliability filter, each stream retained between 2,415 and 4,834 voxels per participant.

\subsection{THINGS Ventral Stream Spiking Dataset (TVSD)}\label{sec:tvsd}

We used the released normalized multiunit activity from monkeys F and N in V1, V4, and IT. Each channel is a multiunit site, not an isolated neuron. The data comprise 22,248 images shown once and 100 images shown 30 times. Electrode reliability was the mean Pearson correlation across all pairs of repetitions on the 100 image set,

\[
\mathrm{Rel}_v = \binom{30}{2}^{-1}\sum_{r<r'} \operatorname{corr}_i\big(y_{irv},\, y_{ir'v}\big),
\]

and electrodes with \(\mathrm{Rel}_v > 0.3\) were retained (F: 332 V1, 157 V4, 117 IT; N: 410, 198, 141). Selection never used model predictions.

For RSA, to evaluate geometry over more than 100 images, the single presentation set (intersected across monkeys and regions) was randomly partitioned: 20\% (4,450 images) for layer selection and 5,000 images sampled from the remainder for reporting.

\subsection{THINGS behavioral similarity}\label{sec:things-behavior}

We used the released 66 dimensional THINGS behavioral embedding for 1,854 object concepts (estimated by the original authors from over 4.70 million triplet odd one out judgments) as a fixed target, with 26,107 associated images. Layer activations were averaged across each concept's images, \(\bar a_{c\ell} = |I_c|^{-1}\sum_{i\in I_c} a_{i\ell}\), and RDMs were computed from these concept vectors. Concepts were randomly split into 20\% for layer selection (370) and 80\% for reporting (1,484); all images follow their concept's assignment. The behavioral RDM uses correlation distance between embedding vectors. Because the embedding is a fixed target with no participant index, bootstrap intervals for this benchmark reflect concept sampling only, not participant or embedding estimation uncertainty.

\section{Feature extraction}\label{sec:feature-extraction}

\subsection{Preprocessing and BatchNorm calibration}\label{sec:preprocessing}

Natural image evaluation used resize to a 256 pixel shorter edge, center crop to 224 $\times$ 224, and ImageNet normalization (CLIP statistics for directly evaluated CLIP models). Inference ran without gradients and with dropout disabled.

For networks with BatchNorm, running statistics were reset and re estimated on evaluation dataset training images that excluded every reporting image (for THINGS, images of selection concepts only). Calibration used one pass over a fixed shuffled order in batches of 64 with cumulative averaging; only BatchNorm modules entered training mode, and no weights or affine parameters were updated. The resulting buffers were fixed for selection and reporting. ResNet-50 was the exception: its BatchNorm statistics were re estimated on a fixed subset of 512,000 ImageNet training images in batches of 256, because coarse trained ResNet-50s have heavy tailed channel activations that the smaller evaluation training sets under sample, and the same statistics were used for both benchmarks. Untrained controls with BatchNorm received the same calibration as their trained counterparts and therefore have random weights but calibrated normalization.

\subsection{Candidate layers}\label{sec:candidate-layers}

Activations were recorded after each nonlinearity (including any BatchNorm), before pooling, and flattened across channels and positions without spatial averaging.

\begin{table}[H]
\centering
\begin{tabular}{ll}
\toprule
Architecture & Candidate layers \\
\midrule
Primary CNN / AlexNet & conv1 to conv5 and fc1, fc2 (post ReLU) \\
ResNet-50 & Stem ReLU; outputs of residual blocks 1, 3, 5, 7, 9, 11, 13, 15, 16 \\
ConvNeXt-Base & Outputs of blocks 3, 6, 9, 14, 19, 24, 29, 33, 36 \\
ViT-B/16 & All 12 encoder block outputs (full token tensor, class plus patch tokens) \\
\bottomrule
\end{tabular}
\end{table}

Classification logits were excluded.
\subsection{Pretrained comparison models}\label{sec:pretrained-models}
The behavioral comparison (Fig.~\ref{fig:behaviour}c) also evaluated off the shelf pretrained networks, one network each, with the same THINGS pipeline (layer selection on the 370 selection concepts, reporting on the 1,484 reporting concepts, concept bootstrap). Supervised ImageNet models (AlexNet, VGG-16, ResNet-50, ViT-B/16, ConvNeXt-Base) used the torchvision ImageNet weights; DINOv1 ResNet-50 the official release via torch hub; DINOv2 ViT-B/14 and DINOv3 ViT-L/16 the timm releases; CLIP ViT-B/32 and ViT-L/14 the OpenAI releases (image encoder only). Candidate layers were the post nonlinearity outputs listed above for the shared architectures; for the others, the block outputs sampled at a regular stride (VGG-16: conv2, 4, 7, 10, 13 and both fully connected layers; DINOv1 ResNet-50: as ResNet-50 plus the projection head; DINOv2: all 12 blocks; DINOv3 and CLIP ViT-B/32: block 1 and every second block; CLIP ViT-L/14: block 1 and every fourth block) plus the output embedding. Preprocessing followed each model's own normalization. These networks differ from the experimental models in data and optimization and serve only as reference points.

\subsection{Sparse random projection}\label{sec:srp}

To bound memory when retaining many layers, features with native dimension \(D > 4{,}096\) were projected to \(k = 4{,}096\) dimensions with a sparse random projection (density \(D^{-1/2}\), nonzero entries \(\pm 1/\sqrt{k\delta}\) with equal probability), generated once and applied identically to selection and reporting images. Features with \(D \le 4{,}096\) were unchanged. In RSA, projected features were used only for layer selection; the selected layer was re extracted at full dimensionality for reporting. In encoding, projected features were used for both selection and the final ridge fit.

\section{Representational similarity analysis}\label{sec:rsa}

RSA compares representational dissimilarity structure across systems\autocite{kriegeskorte2008rsa}. Our inference procedure follows the framework of \textcite{schutt2023}: layers are selected on held out stimuli, scores are reported on an independent partition, and uncertainty is estimated by resampling stimuli with participants held fixed. The full crossvalidated two factor bootstrap for simultaneous generalization to new participants and new stimuli is not implemented.

\subsection{RDMs}\label{sec:rdms}

For a representation matrix \(A \in \mathbb{R}^{N\times D}\) (model layer, ROI response vectors, or behavioral embeddings), the RDM is correlation distance across features,

\[
D^{A}_{ij} = 1 - \operatorname{corr}_d\big(A_{id}, A_{jd}\big),
\]

with correlations clipped to \([-1,1]\) and a zero diagonal. Centering is within each stimulus pattern, not across stimuli. Distances are ordinary correlation distances: not crossvalidated Mahalanobis, not noise whitened, and not bias corrected.

\subsection{Comparison}\label{sec:comparison}

The upper triangles of model and target RDMs (\(N(N-1)/2\) pairs) were compared by Spearman rank correlation with average ranks for ties. Pearson correlation thus defines within space dissimilarities and Spearman correlation compares the two dissimilarity structures.

\subsection{Layer selection on independent stimuli}\label{sec:layer-selection}

For NSD, up to 1,000 selection images were sampled without replacement from each participant's nonshared set with a fixed seed (42); TVSD used its full 4,450 image selection partition; THINGS used its 370 selection concepts. For each candidate layer, selection RSA scores were averaged across subjects with equal weight, and one layer per ROI, model, and training seed was chosen by

\[
\ell^*_r = \arg\max_\ell \frac{1}{S}\sum_{s=1}^{S}\rho^{\text{selection}}_{sr\ell}.
\]

The selected layer is shared across subjects within an ROI and may differ across ROIs and models. Reporting RDMs were then computed from full dimensional activations on reporting images and compared with each subject's RDM (or the behavioral RDM). No reselection occurred on reporting scores. This procedure tests transfer to unseen stimuli within the evaluated subjects; it is not a leave one subject out procedure.

\subsection{Stimulus bootstrap}\label{sec:stimulus-bootstrap}

Uncertainty was estimated with 1,000 resamples drawing \(N\) stimuli (concepts for THINGS) with replacement, with the same draw indexing both RDMs. Pairs of positions referring to the same original stimulus were removed before correlation so that duplicated zero self dissimilarities could not inflate agreement; pairs of distinct stimuli kept their multiplicity. The 95\% interval is the 2.5th and 97.5th percentiles; the point estimate is the full sample score. Layers and weights were fixed across resamples. These intervals capture stimulus sampling uncertainty conditional on the trained networks, selected layers, measured participants, and preprocessing; they do not include uncertainty from network training or layer selection.

\section{Encoding models}\label{sec:encoding}

\subsection{Framework}\label{sec:encoding-framework}

We followed the voxelwise encoding model framework of the Gallant laboratory\autocite{dupre2025}: standardize features and targets, fit ridge regression per target with cross validated regularization, and score predictions on held out images. Fits used one layer at a time with one penalty per target (no banded ridge across feature spaces). Targets are NSD voxels. Because targets are image level response estimates, no temporal delays or HRF fitting were used.

\subsection{Standardization and ridge fit}\label{sec:ridge-fit}

Each feature and target was standardized using statistics from the fitting partition,

\[
\widetilde X_{ij} = \frac{X_{ij} - \mu_j^{\text{fit}}}{s_j^{\text{fit}} + 10^{-8}},
\]

and the same statistics were applied to validation and reporting data. For each target \(v\), ridge regression minimized \(\|\widetilde y_v - \widetilde X w_v\|_2^2 + \alpha_v\|w_v\|_2^2\) without an intercept. The penalty was selected per target by five fold cross validation over 20 logarithmically spaced values from \(10^2\) to \(10^8\), using squared error loss. Fits used the Himalaya GPU ridge regression library\autocite{dupre2022}.

\subsection{Layer selection and reporting}\label{sec:encoding-selection}

For each subject, encoding training images were split 80\% / 20\% into fitting and validation subsets. For each candidate layer, a ridge model was fit on the fitting subset (with the penalty search contained within it) and evaluated on the validation subset; a layer's ROI score was the mean across targets of the Pearson correlation between predicted and observed responses. Validation scores were averaged across subjects and one layer per ROI was selected. The model was then refit on the subject's complete encoding training set, including a new penalty search, and predictions were generated once for the reporting images. The reported score is the arithmetic mean over targets of the held out Pearson correlation,

\[
r_v = \operatorname{corr}_{i\in\text{report}}\big(Y_{iv}, \widehat Y_{iv}\big),\qquad
E_{sr} = \frac{1}{V_{sr}}\sum_{v} r_v,
\]

with no Fisher transform, no reliability weighting, negative correlations retained, and no noise ceiling normalization. For the encoding bootstrap, reporting images were resampled 1,000 times with fixed predictions; correlations were recomputed per target, averaged within ROI, and summarized by 2.5th and 97.5th percentiles.

\section{Aggregation and statistics}\label{sec:statistics}

\subsection{Aggregation}\label{sec:aggregation}

Condition estimates are means across subjects and training seeds with equal weight. Condition level bootstrap distributions are formed by averaging matched resample indices across runs, which is valid because all runs share the same stimulus list, order, and resampling seed.

\subsection{Paired comparisons with the 1,000 class baseline}\label{sec:paired-comparisons}

Paired bootstrap differences use matching resample indices, \(\Delta^{(b)} = \bar\rho^{(b)}_{\text{coarse}} - \bar\rho^{(b)}_{1000}\), summarized by a percentile interval and a two sided tail probability

\[
p_{\text{boot}} = \min\!\left(1,\ \frac{2\,[\,1 + \min(\#\{\Delta^{(b)}>0\},\, \#\{\Delta^{(b)}<0\})\,]}{B+1}\right),
\]

whose minimum attainable value for \(B = 1{,}000\) is \(2/1001\). For NSD, a complementary across participant test averaged scores over seeds within each participant and applied a two sided paired t test across the eight participants; seeds were never counted as additional participants. This test requires at least three subjects and is therefore not applied to TVSD. It is reported in the text rather than in the figures: every comparison marked significant by the stimulus bootstrap in Fig.~\ref{fig:neural} and in the NSD panels of Fig.~\ref{fig:architectures} is also significant under the participant test, with the same sign.

\subsection{Significance testing and multiple comparisons correction}\label{sec:multiplicity}

Within each dataset and ROI panel, Benjamini and Hochberg adjustment at 0.05 was applied separately to the bootstrap and participant test p values across the 18 comparisons (three label sources $\times$ six class counts). A significance marker requires the bootstrap p value to survive adjustment; the same rule applies to both neural datasets. Star thresholds are \(p < 0.05\), \(0.01\), and \(0.0025\) on the unadjusted bootstrap p value; the third threshold reflects bootstrap resolution, which is why it is 0.0025 rather than 0.002: with 1,000 bootstrap iterations and the add-one estimator the smallest attainable p value is 2/1001 = 0.002 and the next is 4/1001 = 0.004. The same rule defines families of 24 comparisons in \edfig{2} (four label sources) and 12 in \edfig{6} (two label sources). Panels with a single label source (Fig.~\ref{fig:architectures}; \edfig{3}) form a family of six comparisons, and the four contrasts of Fig.~\ref{fig:semantic} (handcrafted versus CLIP-derived and 1,000 class labels on two benchmarks) form one family. For the THINGS benchmark, which has no participant index, markers rest on the stimulus bootstrap alone. Differences among coarse class counts, and between experimental models and the pretrained reference models, were not tested and are described qualitatively.

The stimulus bootstrap is conditional on the measured participants, and the participant test on the sampled stimuli; they do not constitute a joint two factor test.

\section{Additional analyses}\label{sec:additional-analyses}

\subsection{Reconstruction from leading principal components}\label{sec:reconstruction}

To ask which variance directions carry representational correspondence, a layer's activation matrix was centered, PCA was fit, the top \(k\) components were retained, and activations were reconstructed in the original feature space,

\[
A^{(k)} = (A - \mathbf{1}\mu^\top) V_k V_k^\top + \mathbf{1}\mu^\top,
\]

before computing correlation distance RDMs and RSA. For the coarse versus 1,000 class comparison, both models were evaluated at the layer selected for the coarse model and \(k\) was swept over \(\{1,2,3,4,5,6,7,8,9,10,12,15,20,25,30,40,50\}\). PCA was fit on the reporting activations, an unsupervised operation on that image set.

\subsection{Per concept alignment}\label{sec:per-concept}

For each THINGS concept \(c\), alignment was the Spearman correlation between its row of the model RDM and its row of the behavioral RDM, excluding the self comparison,

\[
\rho_c = \operatorname{Spearman}\big(D^{\text{model}}_{c,-c},\, D^{\text{behavior}}_{c,-c}\big).
\]

Differences between models identify concepts whose relations to the remaining set are better captured by one representation. Row scores share stimuli and are statistically dependent; their mean does not equal the full triangle RSA. RDM displays order concepts by super-category and, within each, by average linkage hierarchical clustering of the behavioral RDM, with identical ordering across panels; distances are rank transformed for visualization only.

\subsection{PCA visualizations}\label{sec:pca-visualizations}

Two dimensional PCA displays illustrate label partitions in source feature space and the geometry learned by trained networks. Each panel fits its own principal axes; PC signs may be flipped to align centroid arrangements across panels, which resolves PCA's arbitrary orientation and has no effect on any alignment score. Visual separation in two components is descriptive and does not summarize the full geometry.

\clearpage
% Extended Data. Body only: included by main.tex after Methods.
% Figures are placed in order where they appear, sharing a page when they fit; numbered "Extended Data Fig. N | ...".
\makeatletter
\def\fps@figure{H}
\makeatother
\setcounter{figure}{0}
\renewcommand{\figurename}{Extended Data Fig.}
\captionsetup{labelfont=bf, labelsep=bar, font=small}

\section{\Large Extended Data}\label{sec:extended-data}

\begin{figure}
\centering
\includegraphics[width=0.75\linewidth,keepaspectratio]{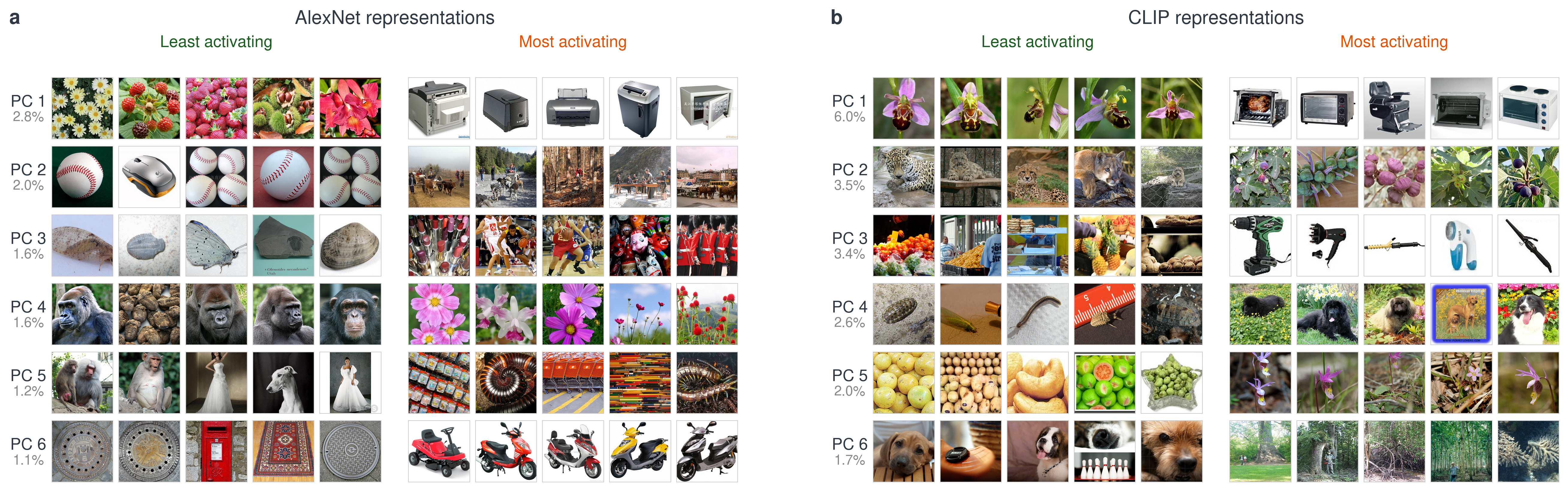}
\caption{\textbf{The image axes used to build the coarse labels.}
For AlexNet (left) and CLIP (right), the five ImageNet training images with the lowest and the highest scores on each of the first six principal components (PC1--PC6) of that network's image representation. Each row is labelled with the component and the percentage of variance it explains. These six axes define the coarse labels: PC1 alone gives the 2-class partition, PC1 and PC2 together give the 4-class partition, and so on up to 64 classes. For AlexNet, PC1 separates colourful natural scenes from achromatic manufactured objects. For CLIP, PC1 separates natural objects from manufactured artefacts and PC2 separates large felines from fruit on branches.}\label{fig:ED1}
\end{figure}

\begin{figure}
\centering
\includegraphics[width=0.85\linewidth,keepaspectratio]{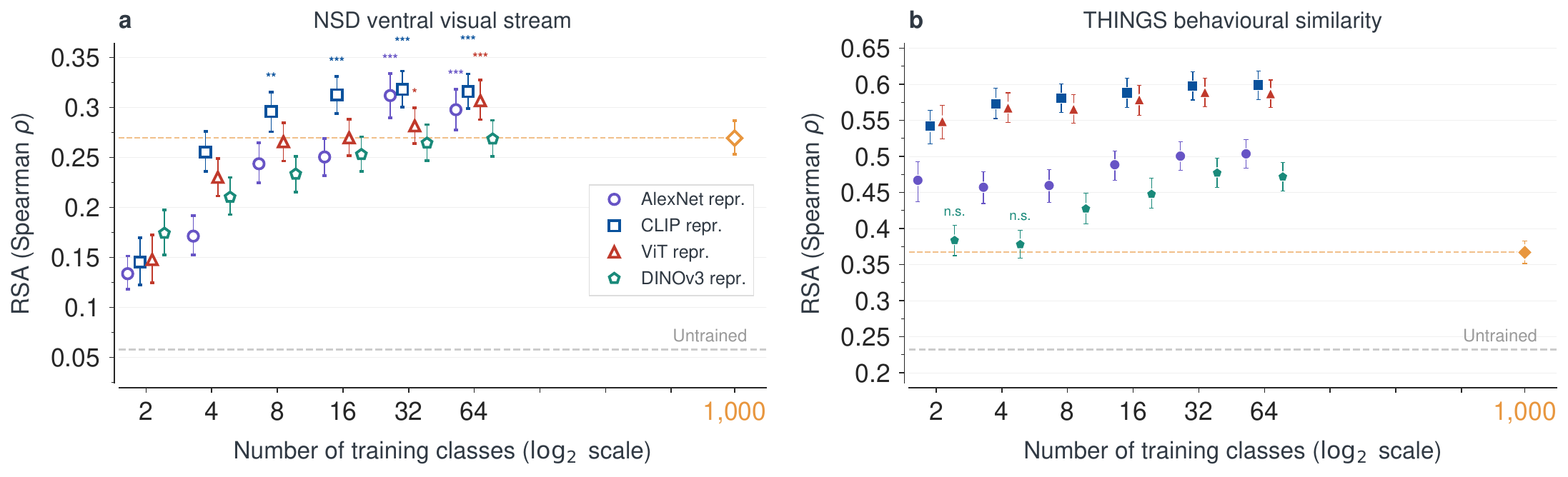}

\caption{\textbf{The benefit of coarse supervision does not depend on which pretrained network defines the labels.}
Alignment of AlexNet-style networks with the human ventral visual stream (\textbf{a}; NSD) and human similarity judgments (\textbf{b}; THINGS) for coarse labels derived from four different source networks: AlexNet (purple circles), CLIP (blue squares), a supervised ViT (red triangles) and DINOv3 (teal pentagons). All networks are trained from scratch on ImageNet using 2--64 classes. The orange diamond and dashed line indicate 1,000-class training; the grey dashed line indicates an untrained network. For every source, coarse training reaches the 1,000-class level in cortex and exceeds it in behavior.
Points are means across three training seeds and, in \textbf{a}, eight participants; error bars are 95\% confidence intervals from 1,000 bootstrap resamples. Significance markers follow Fig.~\ref{fig:neural}. In \textbf{b}, every coarse network exceeds the 1,000-class network except the 2- and 4-class DINOv3-derived networks, marked n.s.}\label{fig:ED2}
\end{figure}

\begin{figure}
\centering
\includegraphics[width=\linewidth,keepaspectratio]{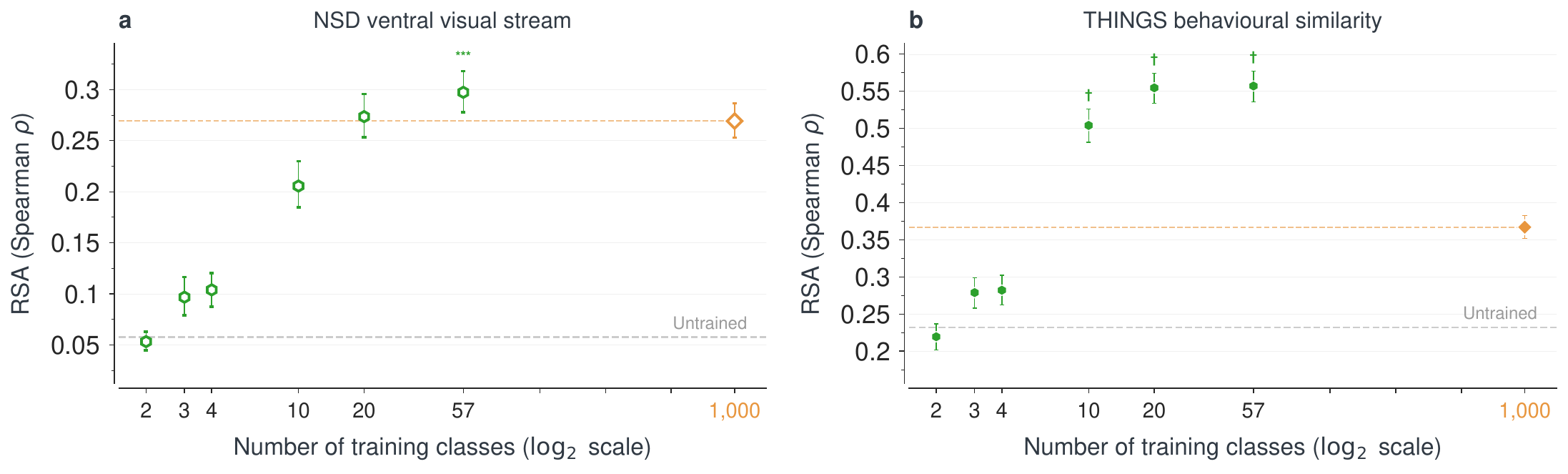}

\caption{\textbf{Coarse labels taken from the WordNet hierarchy also yield high alignment.}
Alignment of AlexNet-style networks with the human ventral visual stream (\textbf{a}; NSD) and human similarity judgments (\textbf{b}; THINGS) when the coarse classes come from the WordNet noun hierarchy rather than from a pretrained network. Each ImageNet category is replaced by its ancestor at a fixed depth of the hierarchy, giving 2, 3, 4, 10, 20 or 57 classes (green hexagons). The orange diamond and dashed line indicate 1,000-class training; the grey dashed line indicates an untrained network. Alignment rises with the number of WordNet classes, exceeding the 1,000-class network in cortex at 57 classes and in behavior from 10 classes onwards.
One network was trained per condition. Error bars are 95\% confidence intervals from 1,000 bootstrap resamples, averaged across eight participants in \textbf{a}. Significance markers follow Fig.~\ref{fig:neural}.}\label{fig:ED3}
\end{figure}

\begin{figure}
\centering
\includegraphics[width=\linewidth,keepaspectratio]{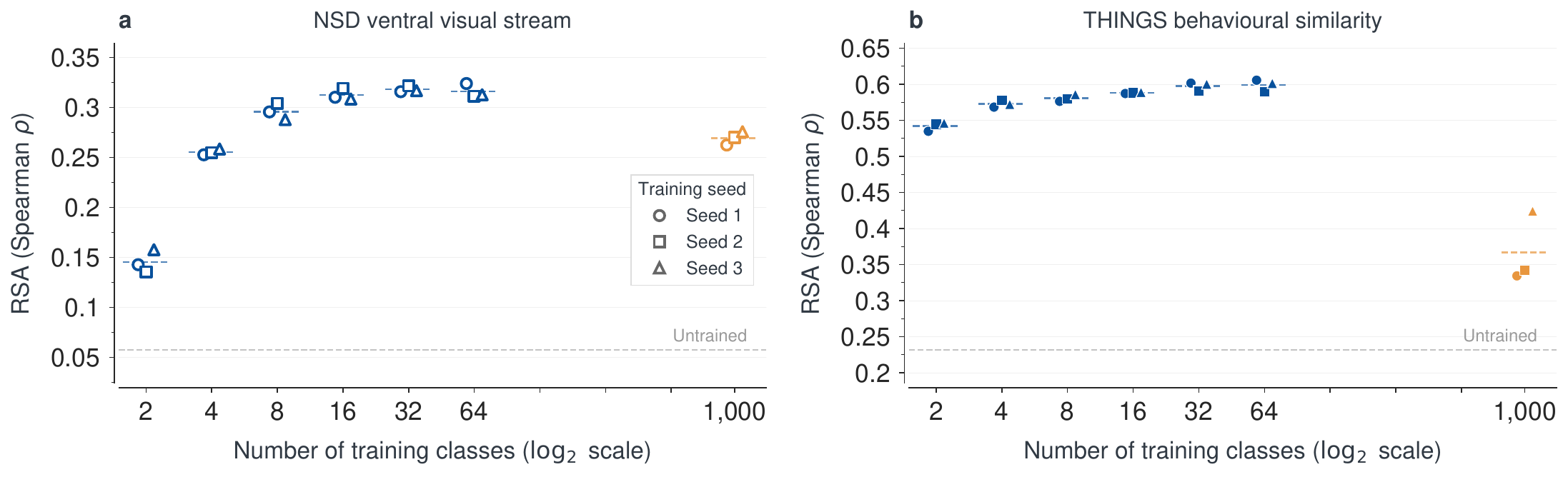}

\caption{\textbf{The results are stable across training seeds.}
Alignment of individual AlexNet-style networks with the human ventral visual stream (\textbf{a}; NSD) and human similarity judgments (\textbf{b}; THINGS) for 2--64 CLIP-derived classes (blue) and 1,000 classes (orange). Each condition was trained three times from different random initializations (circle, seed 1; square, seed 2; triangle, seed 3); a short dashed tick marks their mean. The grey dashed line indicates an untrained network. The three networks in each condition lie close together, and every coarse network exceeds every 1,000-class network in \textbf{b}. The 1,000-class networks spread more widely in \textbf{b} because one seed selects a later layer. Vertical ranges match Fig.~\ref{fig:neural}d and Fig.~\ref{fig:behaviour}b.}\label{fig:ED4}
\end{figure}

\begin{figure}
\centering
\includegraphics[width=0.5\linewidth,keepaspectratio]{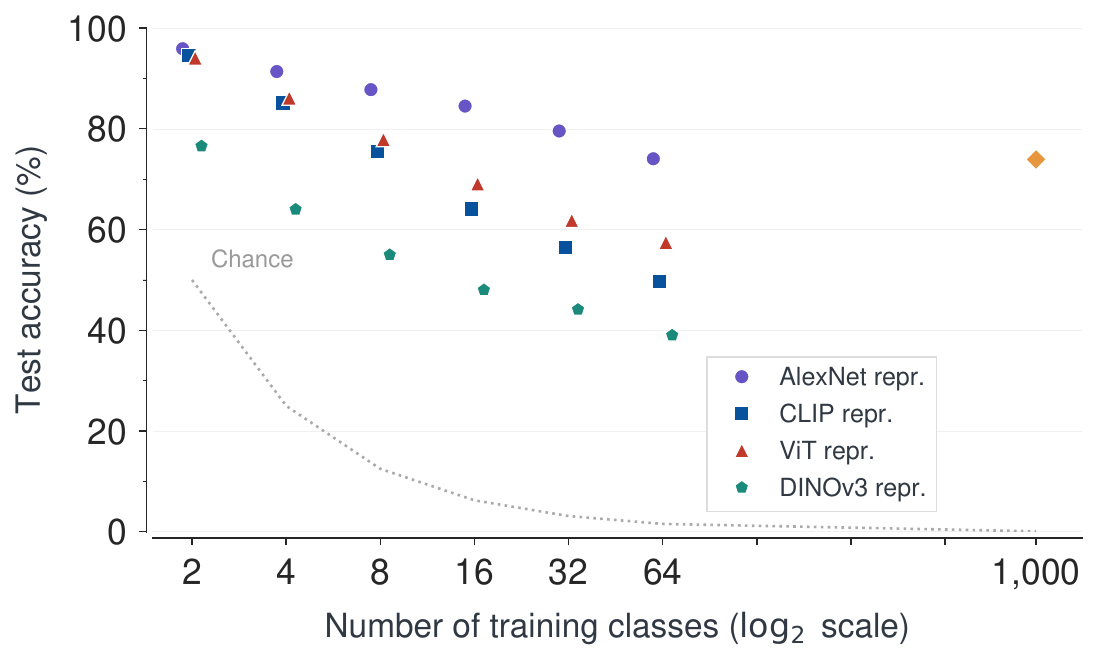}
\caption{\textbf{All networks learn their task, but accuracy does not track alignment.}
Held-out classification accuracy of AlexNet-style networks trained on 2--64 coarse classes derived from AlexNet (purple circles), CLIP (blue squares), a supervised ViT (red triangles) or DINOv3 (teal pentagons), and on the 1,000 ImageNet categories (orange diamond). The dotted grey curve shows chance accuracy. Every network performs far above chance. Accuracy falls as the number of classes grows and differs widely between label sources at the same number of classes, yet the four sources yield similar alignment (\edfig{2}). Points are means across three training seeds; the standard error is smaller than the markers.}\label{fig:ED5}
\end{figure}

\begin{figure}
\centering
\includegraphics[width=0.9\linewidth,keepaspectratio]{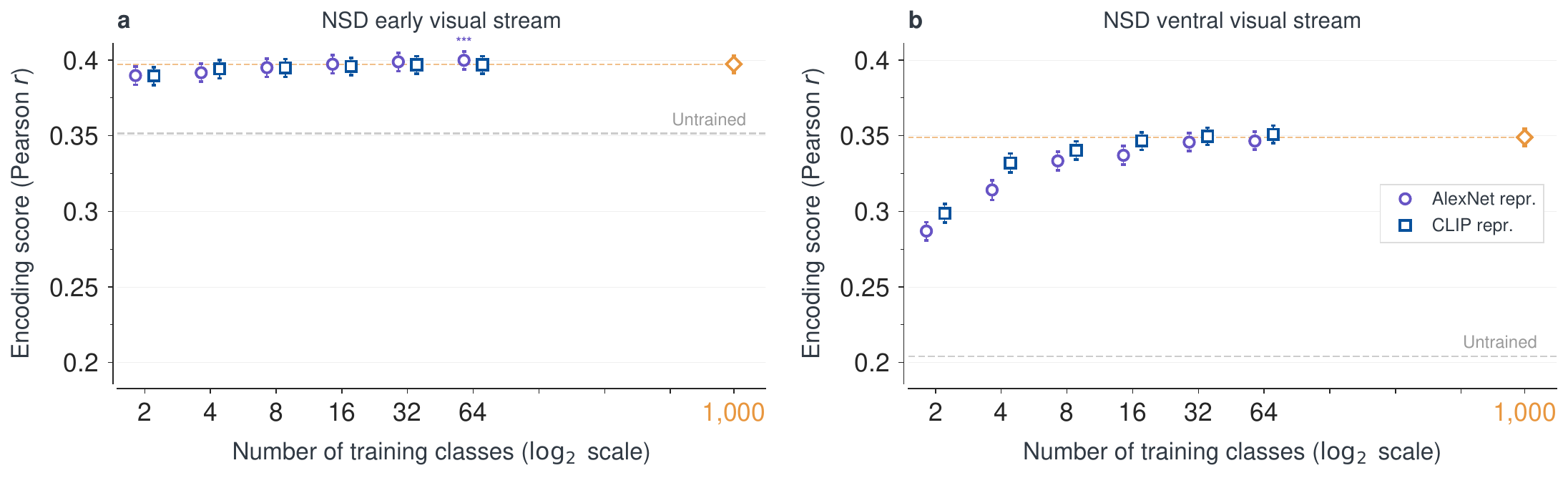}
\caption{\textbf{Encoding models show the same pattern as RSA.}
Alignment with the human early visual stream (\textbf{a}) and ventral visual stream (\textbf{b}; NSD) measured with voxelwise encoding models rather than RSA. Network features are linearly reweighted to predict each voxel's response, and the score is the mean correlation between predicted and measured responses on held-out images. AlexNet-style networks are trained on 2--64 classes derived from AlexNet (purple circles) or CLIP (blue squares); the orange diamond and dashed line indicate 1,000-class training; the grey dashed line indicates an untrained network. As with RSA, coarse-trained networks approach the 1,000-class level in the ventral stream from 8 classes and match it by 32, and training has little effect in the early visual stream.
Points are means across three training seeds and eight participants; error bars are 95\% confidence intervals from 1,000 bootstrap resamples. Significance markers follow Fig.~\ref{fig:neural}.}\label{fig:ED6}
\end{figure}

\begin{figure}
\centering
\includegraphics[width=\linewidth,keepaspectratio]{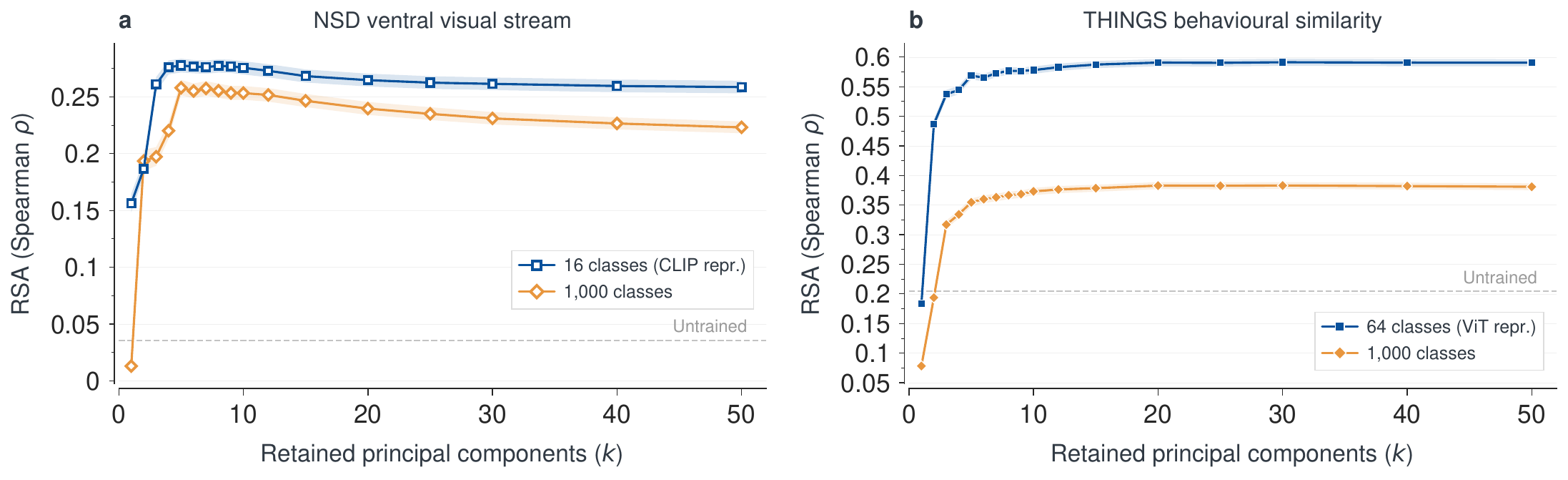}

\caption{\textbf{A low-dimensional version of the 1,000-class representation does not recover the alignment of a coarse-trained network.}
Alignment with the human ventral visual stream (\textbf{a}; NSD) and human similarity judgments (\textbf{b}; THINGS) when each network's representation is reduced to its first $k$ principal components before comparison. A coarse-trained network (blue squares; 16 CLIP-derived classes in \textbf{a}, 64 ViT-derived classes in \textbf{b}) is compared with the 1,000-class network (orange diamonds). Both are AlexNet-style networks; the grey dashed line indicates an untrained network. If the coarse network's representation were simply a low-dimensional subspace of the 1,000-class representation, the two curves would converge at small $k$. Instead, the coarse network is more aligned from $k = 3$ onwards, and the gap persists as more components are retained. Bands are 95\% confidence intervals from 1,000 bootstrap resamples.}\label{fig:ED7}
\end{figure}

\begin{figure}
\centering
\includegraphics[width=\linewidth,keepaspectratio]{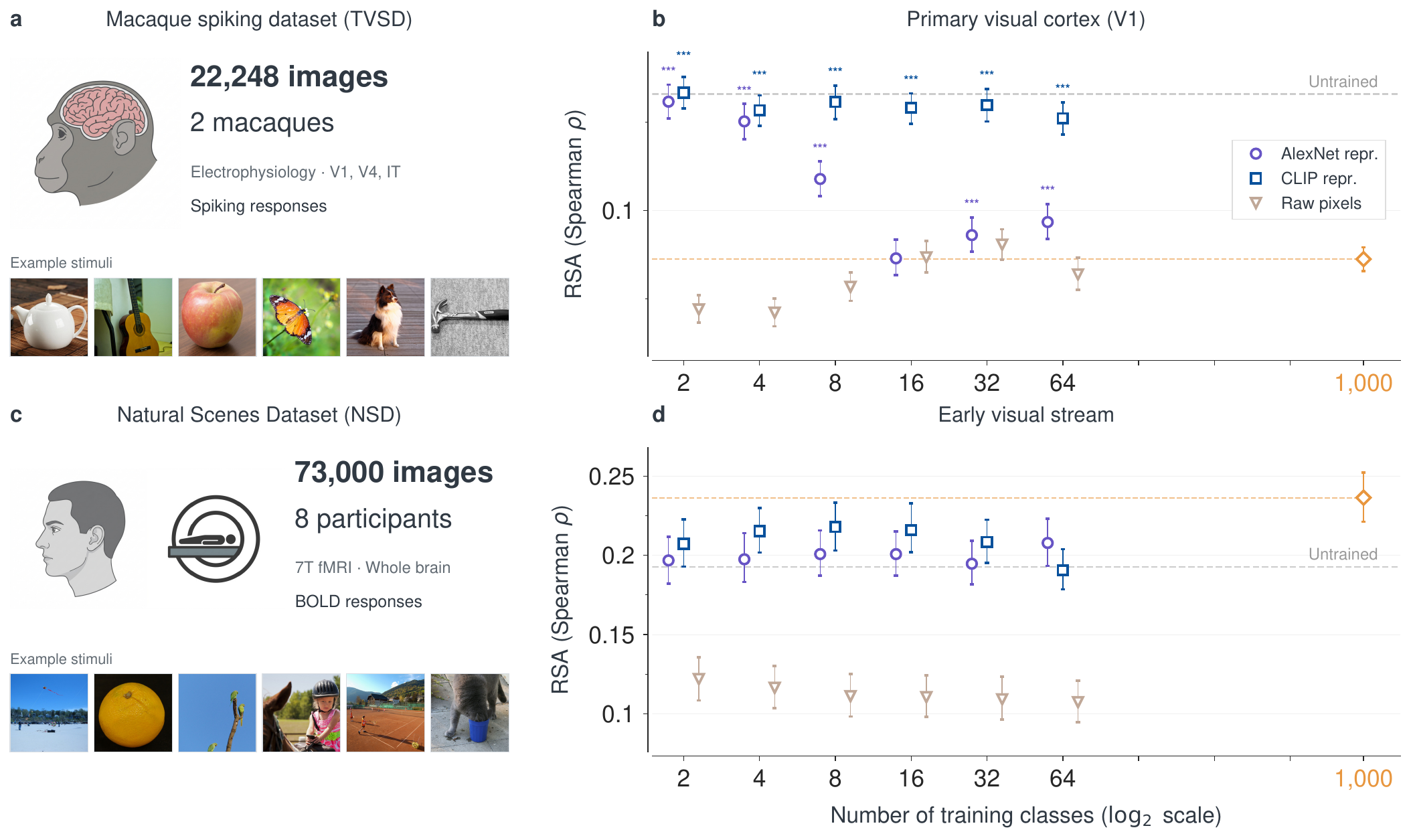}
\caption{\textbf{In early visual cortex, alignment depends little on training.}
The early-visual counterpart of Fig.~\ref{fig:neural}, using the same datasets and conventions.
\textbf{a,c}, The macaque spiking dataset (TVSD; \textbf{a}) and the human fMRI dataset (NSD; \textbf{c}).
\textbf{b,d}, Alignment with macaque primary visual cortex (V1; \textbf{b}) and the human early visual stream (\textbf{d}) for AlexNet-style networks trained on 2--64 coarse classes derived from AlexNet (purple circles), CLIP (blue squares) or raw pixels (tan triangles). The orange diamond and dashed line indicate 1,000-class training; the grey dashed line indicates an untrained network. In contrast to high-level cortex, training changes alignment little here: in macaque V1 the untrained network is the most aligned model, and in the human early visual stream all trained networks lie within a narrow range of it. This pattern, an untrained network matching or exceeding trained networks in early visual cortex but not in higher areas, has been reported previously\autocite{cichy2016,farahat2025,kazemian2025}. Pixel-derived labels remain below every other condition.
Points are means across individuals and three training seeds; error bars are 95\% confidence intervals from 1,000 bootstrap resamples. Significance markers follow Fig.~\ref{fig:neural}.}\label{fig:ED8}
\end{figure}

\end{document}